\documentclass{article} % For LaTeX2e
\usepackage[dvipsnames]{xcolor}
\usepackage{iclr2025_conference,times}

\usepackage{amsmath,amsfonts,bm}

\def\eqref#1{equation~\ref{#1}}
\def\1{\bm{1}}

\def\vs{{\bm{s}}}

\DeclareMathAlphabet{\mathsfit}{\encodingdefault}{\sfdefault}{m}{sl}
\SetMathAlphabet{\mathsfit}{bold}{\encodingdefault}{\sfdefault}{bx}{n}

\DeclareMathOperator{\sign}{sign}

\usepackage{hyperref}
\usepackage{url}
\usepackage{bbm}
\usepackage{cleveref}
\usepackage{caption}

\usepackage{graphicx}
\usepackage{booktabs}
\usepackage{multirow}
\usepackage{xspace}
\usepackage{pifont}% http://ctan.org/pkg/pifont
\usepackage{colortbl} % 导入颜色表格包

\usepackage{enumitem}
\usepackage{wrapfig}

\newcommand{\cmark}{\ding{51}}%
\newcommand{\xmark}{\ding{55}}%

\newcommand{\modelname}{BiGR\xspace}
\newcommand{\rebut}[1]{#1}

\title{\modelname: Harnessing Binary Latent Codes for Image Generation and Improved Visual Representation Capabilities}

\author{Shaozhe Hao$^{1}$ \quad Xuantong Liu$^{2}$ \quad Xianbiao Qi$^{3}$\thanks{Project lead} \quad Shihao Zhao$^{1}$ \quad Bojia Zi$^{4}$ \\
\textbf{Rong Xiao$^{3}$ \quad \quad Kai Han$^{1}$\thanks{Corresponding authors} \quad \quad Kwan-Yee K. Wong$^{1}$}\footnotemark[2] \\
$^1$The University of Hong Kong \quad $^2$Hong Kong University of Science and Technology \\ 
$^3$Intellifusion \quad $^4$The Chinese University of Hong Kong \\
\texttt{\small \{szhao,shzhao,kykwong\}@cs.hku.hk qixianbiao@gmail.com kaihanx@hku.hk} \vspace{3pt} \\
Project page: \url{https://haoosz.github.io/BiGR} \\
Code and models: \url{https://github.com/haoosz/BiGR}
}

\hypersetup{
    colorlinks=true,
    linkcolor=BrickRed,
    filecolor=cyan,      
    urlcolor=magenta,
    citecolor=Blue,
}

\crefname{section}{sec.}{sec.}
\Crefname{section}{Sec.}{Sec.}
\crefname{figure}{fig.}{fig.}
\Crefname{figure}{Fig.}{Fig.}
\crefname{equation}{eq.}{eq.}
\Crefname{equation}{Eq.}{Eq.}
\crefname{table}{tab.}{tab.}
\Crefname{table}{Tab.}{Tab.}

\def\eg{\emph{e.g.}}

\def\ie{\emph{i.e.}}

\def\vs{\emph{vs.}}

\iclrfinalcopy % Uncomment for camera-ready version, but NOT for submission.

\begin{document}

\maketitle

\vspace{-15pt}
\begin{figure}[ht]
    \centering
    \includegraphics[width=1.0\linewidth]{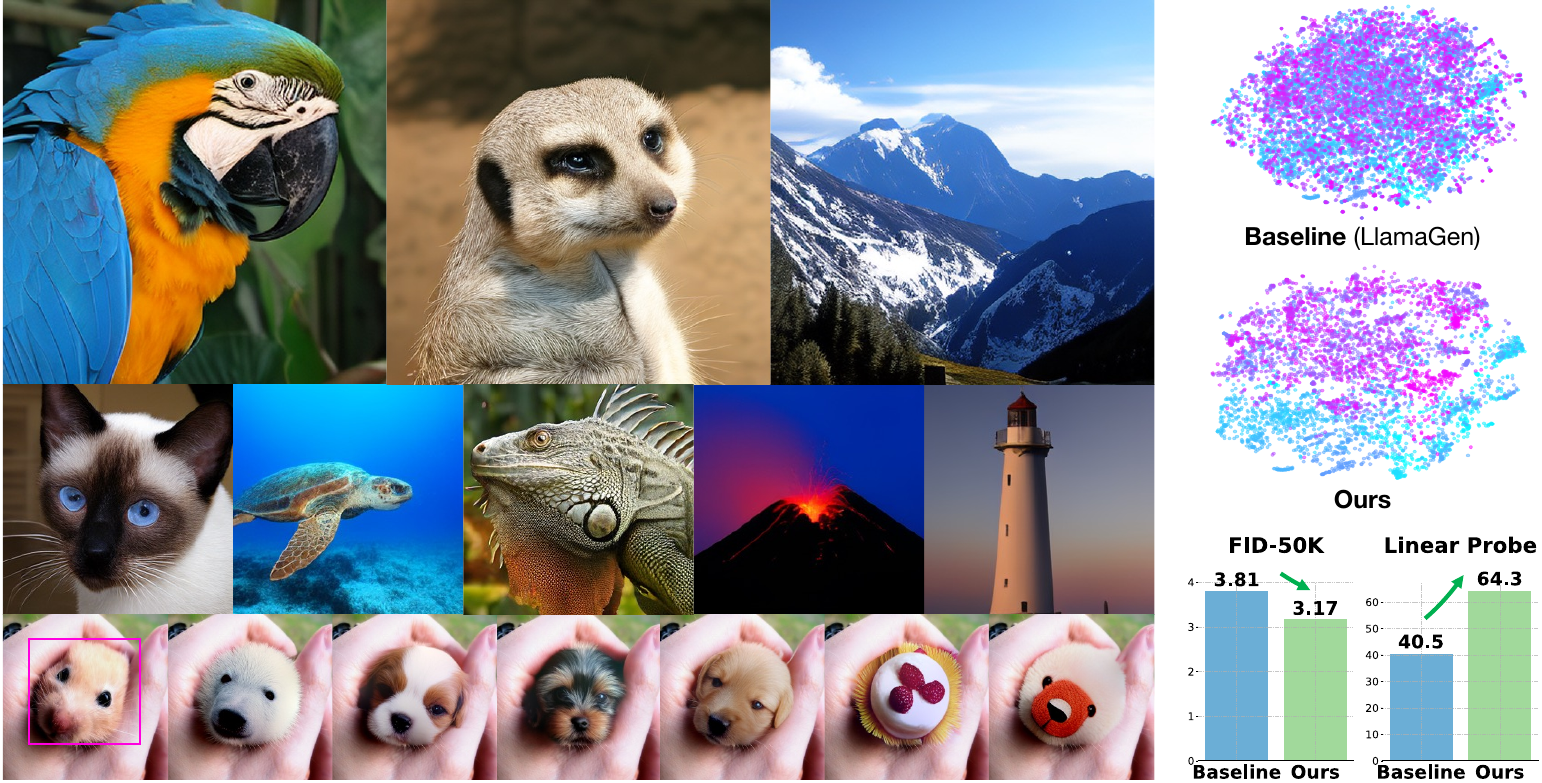}
    \caption{
    \textbf{\modelname generates high-quality images while improving the discriminative capabilities of the representations.} 
    \textbf{Left}: Generated 512$\times$512 samples, 256$\times$256 samples, and class-conditional editing samples. \textbf{Right}: \modelname \vs\   LlamaGen~\citep{sun2024autoregressive}.
    We visualize image features from 100 classes in ImageNet-1K validation split using t-SNE~\citep{tsne}, where the same color indicates the same class.
    Our model produces features with greater discriminative separability and enhances both generative and discriminative performance.
    }
    \label{fig:teaser}
\end{figure}

\begin{abstract}
\vspace{-8pt}
We introduce \modelname, a novel conditional image generation model using compact binary latent codes for generative training, focusing on enhancing both generation and representation capabilities. \modelname is the first conditional generative model that unifies generation and discrimination within the same framework. 
\modelname features a binary tokenizer, a masked modeling mechanism, and a binary transcoder for binary code prediction. 
Additionally, we introduce a novel entropy-ordered sampling method to enable efficient image generation. 
Extensive experiments validate \modelname's superior performance in generation quality, as measured by FID-50k, and representation capabilities, as evidenced by linear-probe accuracy. 
Moreover, \modelname showcases zero-shot generalization across various vision tasks, enabling applications such as image inpainting, outpainting, editing, interpolation, and enrichment, without the need for structural modifications. Our findings suggest that \modelname unifies generative and discriminative tasks effectively, paving the way for further advancements in the field. \rebut{We further enable \modelname to perform text-to-image generation, showcasing its potential for broader applications.}
\end{abstract}

% \newpage

\section{Introduction}
\vspace{-5pt}
Image generation is experiencing a revolutionary growth driven by the advancements in diffusion models~\citep{ho2020denoising,rombach2022high,peebles2023scalable,ma2024sit} and autoregressive models~\citep{esser2021taming, sun2024autoregressive, tian2024visual}. While these models have demonstrated impressive performance, % in image generation, 
their representation capabilities are under-studied. As revealed by~\citet{balestriero2024how}, reconstruction-based learning often produces visually compelling results but fails to provide strong latent representations for perception. It has been a long-desired goal of the research community to design a good image generator which can also serve as a strong feature extractor.

Centered around this goal, previous studies~\citep{chen2020generative,li2023mage} on representation capabilities of generative models have primarily focused on unconditional generation. 
Despite conditional generation~\citep{peebles2023scalable,sun2024autoregressive,li2024autoregressive} has emerged as a recent research trend and garnered much attention, investigations of the representation capabilities of conditional generative models remain limited.
In conditional image generation, conditions are added to guide the generation process. However, this guidance is absent in downstream discriminative tasks. This weakens the relationship between features and categories, and thereby diminishes the representation capabilities of the features.
We validate this limitation using the latest class-conditional image generation model~\citep{sun2024autoregressive} (see~\Cref{fig:teaser} (right)), and stress the necessity of improving the representation capabilities of conditional generative models.

We introduce \textbf{\modelname}, a novel conditional image generation model that utilizes compact \textbf{Bi}nary latent codes for \textbf{G}enerative tasks with improved \textbf{R}epresentation capabilities.
\modelname is trained exclusively through a generative process by reconstructing tokens without relying on any discriminative losses. 
We compress an image into a sequence of binary codes using lookup-free quantization~\citep{yu2024language,wang2023binary} and utilize our model to predict these binary codes.
We emphasize that \modelname is the first conditional image generation model that unifies generative and discriminative tasks, achieving improved performance across both.
Below, we describe our model design, generative and discriminative use, and zero-shot generalized applications.

Our framework, built upon the language model architecture, has three major components, namely \textbf{(1)} a binary tokenizer that converts a pixel-level image into a sequence of binary latent codes, \textbf{(2)} a transformer equipped with full bidirectional attention, and \textbf{(3)} a binary transcoder that transforms continuous features into Bernoulli-distributed binary codes. 
We train \modelname using the masked modeling approach~\citep{bao2022beit,chang2022maskgit,li2023mage}. 
This modification, deviating from the typical autoregressive approach, expands token interaction without altering the structure of Llama.
Paired with a tailored inference process and inherent visual representations, \modelname can perform both generative and discriminative tasks. 

For \textit{generative} purpose, we design a sampling method that iteratively unmask tokens in a sequence, with the order determined by the binary entropy magnitude from the predicted Bernoulli distribution probabilities. 
This approach requires only a small number of sampling iterations which significantly accelerates the inference process.
As a result, we achieve high efficiency in image generation compared with diffusion models, which require multiple steps to remove noise. 
% and autoregressive models, which predict each token sequentially. 
Through extensive experiments, we show that \modelname performs on par with, or even surpasses, existing baselines in quantitative metrics.

For \textit{discriminative} purpose, we perform average pooling on the intermediate features in \modelname. By this straightforward operation, \modelname exhibits significantly stronger representation capabilities than comparable models, which has been empirically validated through linear probe evaluation.
Due to the compactness of binary codes and the global information from masked modeling, the feature representations produced by \modelname can more effectively linearly separate visual categories in downstream discriminative tasks.

Moreover, we explore the zero-shot generalization capabilities of \modelname within the generation domain.
Unlike autoregressive models that must predict tokens in raster order, the masked modeling mechanism offers a huge flexibility during inference, allowing for the design of task-specific strategies. 
As a result, \modelname can perform various vision tasks in a zero-shot manner, without requiring any structural changes or parameter fine-tuning. 
In this paper, we showcase applications of our model in image inpainting, outpainting, editing, interpolation, and enrichment.
\rebut{We further extend \modelname to perform text-to-image generation, highlighting its potential for broader applications. The generated results are provided in the Appendix.}
We believe that further applications of \modelname can be unlocked through community efforts.
% make sure image enrichment does not denote other things.

% The key breakthroughs of \modelname are as follows:
To summarize, our \modelname possesses the following prominent advantages:
% \begin{enumerate}[leftmargin=*]
% \item 
\textbf{(i) Uniformity} - \modelname is the first conditional image generation model that unifies generative and discriminative tasks within the same model. By modeling compact binary latent codes, \modelname delivers strong performance in both tasks compared to existing models.
\textbf{(ii) Efficiency} - \modelname generates images at a low time cost, attributed to the small number of sampling steps required in the iterative unmasking process, while still maintaining high generation quality.
\textbf{(iii) Flexibility} - \modelname can be flexibly employed for various vision applications, such as inpainting, outpainting, editing, interpolation, and enrichment in a zero-shot manner, without the need for task-specific structural changes or parameter fine-tuning.
\textbf{(iv) Scalability} - \modelname demonstrates scalability in both generative and discriminative tasks, as evidenced by comprehensive evaluations of both generation quality and linear-probe performance.
% \end{enumerate}
\vspace{-2pt}
\section{Related work}
\vspace{-2pt}
\textbf{Binary latent code modeling}\quad
Binary latent code, also known as hashing~\citep{wang2017survey}, has been largely demonstrated effective for visual representations due to its compactness and discreteness~\citep{cakir2019hashing, jiang2018asymmetric, shen2015supervised, wei20212,wu2019deep}. In the realm of visual generation, the study of binary tokenizers has recently attracted notable attention, referred as look-up free quantization in~\citet{yu2024language} and as binary autoencoder in~\citet{wang2023binary}. Binary tokenizers can enhance the codebook utilization for vector-quantization methods~\citep{esser2021taming,van2017neural}, facilitating image and video generation. \citet{wang2023binary} introduces a Bernoulli diffusion process that operates on Bernoulli-distributed variables to generate binary latents. Our work studies this type of tokenizers and we propose a novel generative framework for uniform conditional generation and visual representation.

\textbf{Generative representation learning}\quad
Representation learning has long been an important topic, with self-supervised methods~\citep{he2020moco, chen2020simclr, caron2020swav, grill2020byol, caron2021dino, zhou2022ibot} dominating the field in the past few years. Some works learn visual representations through generative modeling. For example, iGPT~\citep{chen2020generative} predicts pixels in a manner similar to GPTs~\citep{gpt3}, while MAE~\citep{he2022masked} and MAGE~\citep{li2023mage} reconstruct masked image regions.
ViT-VQGAN~\citep{yu2022vector} studies the representation capabilities of unsupervised generative models.
However, these methods involve specialized designs for discriminative tasks and are not directly suited for conditional image generation. Our work broadens this scope by proposing a conditional image generation framework that consistently delivers both high-quality generation and strong representation capabilities.

\textbf{Conditional image generation}\quad
Conditional image generation has gained significant attention recently. Existing works on this topic can be broadly grouped into two categories: \textit{diffusion} models~\citep{ho2020denoising, songdenoising, rombach2022high, peebles2023scalable, ma2024sit, chen2024pixartalpha}, which gradually denoise a random Gaussian noise, and \textit{autoregressive} models~\citep{esser2021taming, yu2022scaling, yu2022vector, sun2024autoregressive, tian2024visual}, which predict the next tokens similarly to language models.
The models based on masked prediction~\citep{chang2022maskgit, li2023mage, chang2023muse} can be classified as autoregressive models, as discussed in~\citep{li2024autoregressive}.
In this paper, for clarity, we use ``autoregressive'' to specifically refer to models that use causal attention and next-token prediction, and ``mask'' to refer to models using masked modeling.
% It remains a matter of dispute~\citep{kilian2024computational} which type of models offers better performance and scalability. 
Although conditional generative models can produce visually compelling images, their representation capabilities have rarely been studied. Our work aims to bridge this gap.

\vspace{-2pt}
\section{Method}
\begin{figure}
    \centering
    \includegraphics[width=1.0\linewidth]{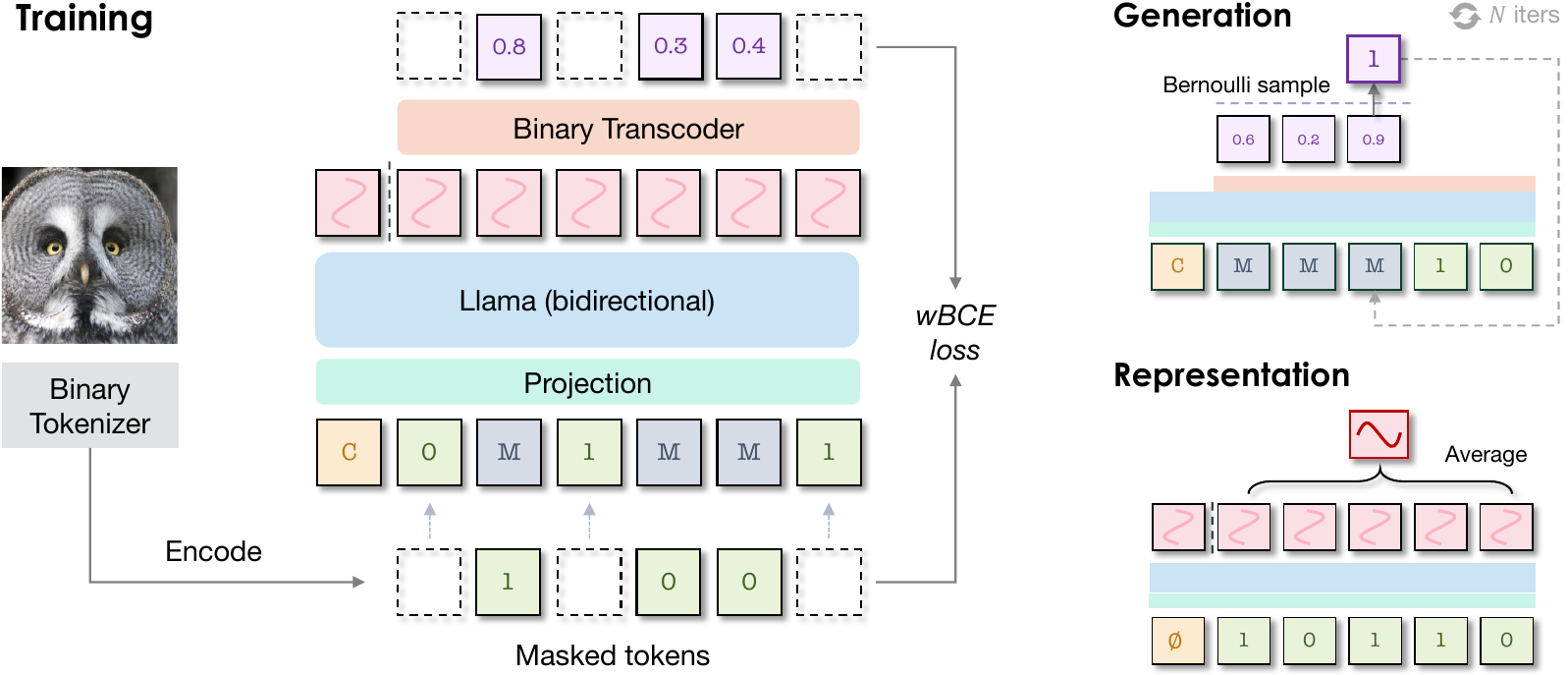}
    \caption{\textbf{Overview of \modelname.} For simplicity, we display only 1 bit for each token, although each token actually consists of K bits in length. \textbf{Left}: We outline the training of \modelname. Starting with binary codes from binary tokenizers, we append a condition token and mask partial tokens. These tokens are projected into continuous embeddings and processed by the Llama backbone. The outputs undergo a Bernoulli denoising process in the binary transcoder to generate probabilities, penalized by the weighted binary cross-entropy loss (wBCE) at masked positions. \textbf{Right}: We illustrate the generation process (detailed in~\Cref{sec:sampling}) and the representation acquisition via average pooling.}
    \label{fig:overview}
\end{figure}

\vspace{-2pt}
Our framework is based on a masked language model that operates directly on binary latent codes derived from images. 
We train the model by masking a portion of the input tokens and learning to unmask them using predicted output tokens. The prediction is achieved through a Bernoulli diffusion process \citep{wang2023binary}, which is well-suited for generating binary latent codes. 
In sampling, we determine the order of tokens to be unmasked based on the magnitude of entropy computed from the predicted Bernoulli distribution probabilities.
To obtain latent representations, we perform average pooling on the intermediate features of our model.
We present the overview of \modelname in~\Cref{fig:overview}.
We describe the details of each of its components below.

\subsection{Preliminary}
We first review the binary tokenizer and the Bernoulli diffusion process that underpin our model.

\textbf{Binary tokenizer}\quad An image tokenizer $\mathcal{T}$ can encode an image $x \in \mathbb{R}^{3\times H \times W}$ into a sequence of latent codes $\{\zeta^1, \zeta^2, \cdots, \zeta^{n}\}=\mathcal{T}(x)$, where each $\zeta^i$ represents the code at a specific spatial position.
Binary tokenizers~\citep{yu2024language,wang2023binary}, also known as lookup-free quantization, transform the code into binary format by
\begin{equation}
    z^i = \sign(\zeta^i) = \mathbbm{1}\{\zeta^i>0\},
\end{equation}
and a corresponding token index $r^i$ can be computed by
\begin{equation}
\label{index_q}
    r^i = \sum^{K}_{k=1} 2^{k-1} \cdot z^i[k],
\end{equation}\vspace{-13pt}

where $z^i[k]$ denotes the $k$-th bit of the binary code $z^i$, and $K$ is the number of binary bits (\ie, code dimension), resulting in a total of $2^K$ token indices.
Using~\Cref{index_q}, \citet{yu2024language} indexes image tokens with the binary code $z^i$ and build a vocabulary of size $2^K$ for generative purposes.
In contrast, our approach focuses on directly modeling the sequence of binary codes $\{z^i\}_{i=1}^n$.

\textbf{Bernoulli diffusion}\quad We generate binary codes through a Bernoulli diffusion process~\citep{wang2023binary}, which effectively models Bernoulli-distributed variables. 
Specifically, Bernoulli diffusion process adds Bernoulli noise from the starting point $z \sim q(z^0)$:
\begin{equation}
    q(z^t | z^{t-1}) = \mathcal{B}\left(z^{t}; z^{t-1}(1-\beta^t) + 0.5\beta^t\right) \quad t = 1, 2, \cdots, T.
\end{equation}
Here, $\mathcal{B}$ denotes a Bernoulli distribution, and the timestep $t$ out of the total $T$ is denoted as a superscript. 
We model the denoising process by $p(z^{t-1} | z^t)$, which predicts the Bernoulli distribution probabilities for the binary code at the previous timestep.
By iterating the denoising process, starting with a coin toss $\mathcal{B}(0.5)$, we can finally generate binary codes that follow Bernoulli distributions. \rebut{In our model, the \textbf{binary transcoder} is the component that integrates the Bernoulli diffusion process, responsible for denoising Bernoulli noise and predicting binary codes.}

\vspace{-2pt}
\subsection{Masked modeling on binary latent codes}
\vspace{-2pt}
\label{sec:mlm-bin}

\textbf{Backbone}\quad 
We build our method upon the transformer-based language model Llama~\citep{dubey2024llama3,touvron2023llama2,touvron2023llama1}. Unlike language, an image is not naturally modeled as a causal sequence of tokens, but instead, each token should have access to all others to better capture global visual information.
Therefore, we replace the causal attention commonly used in language models with bidirectional attention, and let the model predict masked tokens instead of next tokens.

\textbf{Input projection}\quad 
In the input space, instead of looking up an embedding vector with a token index, we use a simple linear layer that projects the binary code onto the embedding space. This technique has recently been explored for continuous-valued tokenizers in~\citet{tschannen2023givt}, and we find that it also works well for binary-valued tokenizers.
We maintain standard conditional embeddings and mask embeddings, where the conditional embedding is appended at the start of the sequence, and the mask embedding replaces inputs at masked positions.

\textbf{Mask-token prediction}\quad
During training, we simply mask a portion of image tokens with a learnable mask token \texttt{[M]}. The fraction of masked tokens follows a cosine schedule, as used in~\citet{li2023mage}. We compute losses only for the masked positions, where the model predicts the values of the masked tokens. Formally, let $f_\theta$ represent the language model, and $\{z_{m_i}^{i}\}_{i=1}^n$ denote the sequence of binary codes that are partially masked. Here, $M=\{m_i\}_{i=1}^n$ indicates whether the $i$-th position is masked ($m_i = 1$) or left unmasked ($m_i = 0$).
We obtain outputs at the masked positions from the language model $\{h^i\}_{m_i=1} = f_\theta(\{z_{m_i}^{i}\}_{i=1}^n)$, which are distributed in a continuous space.

\textbf{Binary transcoder}\quad
We transform the model outputs $h$ into binary codes\footnote{Since the operation is position-wise, we omit the superscript of positions for simplicity.} $z$ through a Bernoulli diffusion process~\citep{wang2023binary}.
In particular, we learn a denoising network $g_\phi$ with a Sigmoid function $S$ to model
\begin{equation}
\label{eq:p_process}
    p_\phi (z^{t-1} | z^t) = \mathcal{B}\left(z^{t-1}; S(g_\phi(z^t, t, h))\right),
\end{equation}
which predicts the probabilities of the Bernoulli distribution conditioned on the intermediate feature $h$. Consequently, binary codes can be generated by sampling from these probabilities. Following~\citet{wang2023binary,ho2020denoising}, the training target is the binary residual, \ie, $z^{t} \oplus z^0$ where $\oplus$ represents the element-wise XOR operation. The training objective is simply an element-wise weighted binary cross-entropy (wBCE) loss, expressed as
\begin{gather}
y_k = (z^{t} \oplus z^0)[k] \in \{0,1\} \quad p_k = S(g_\phi(z^t, t, h))[k] \in [0,1], \\
\label{eq:bin_loss}
\mathcal{L} = -\frac{1}{K}\sum^{K}_{k=1} w_k \left( y_k \log p_k + (1 - y_k) \log(1 - p_k) \right), \\ 
\text{where} \quad w_k = (1-y_k) \cdot \sum^{K}_{k=1} y_k / K + y_k \cdot (1-\sum^{K}_{k=1} y_k / K) + 1/K.
\end{gather}\vspace{-10pt}

Here, $y_k$ represents the target, and $p_k$ is the predicted probability for the $k$-th bit in the binary code.
The element-wise loss weight $w_k$ is applied to mitigate the imbalance between $0$s and $1$s, calculated based on their respective ratios in a $K$-dimensional code. The constant term $1/K$ is added to prevent nearly-zero weights that could impede training. In training, we jointly optimize the language model $f_\theta$ and the denoising network $g_\phi$ using the loss defined in~\Cref{eq:bin_loss}.

\textbf{Visual representation}\quad
Once trained, our model inherently possesses strong visual representations. Given an image, we input it into the model without any masks, along with an unconditional token appended. We then perform average pooling on the continuous-valued features $h$ to derive the global representation of the given image. We observe that the most discriminative representation originates not from the final layer but from the middle layers within the transformer blocks, in line with the findings in~\citet{yu2022vector,chen2020generative}. 
As a result, we use the intermediate features as the final image representation.
\vspace{-2pt}
\subsection{Entropy-ordered generative sampling}
\vspace{-2pt}
\label{sec:sampling}
For image generation, we design a sampling strategy for our model, enabling it to iteratively predict tokens from a fully masked sequence. Unlike in training, where mask positions are randomly chosen at each step, during sampling, the order in which tokens are unmasked follows a predefined criterion.

We arrange the masked tokens according to the binary entropy magnitude calculated from the predicted probabilities. The binary entropy is defined as:
\begin{equation}
\label{eq:bin_ent}
\mathcal{H} = -\frac{1}{K}\sum^{K}_{k=1} p_k \log_2 p_k + (1 - p_k) \log_2 (1 - p_k),
\end{equation}
which ranges from 0 to 1. 
Here, a low value indicates high prediction confidence (\ie, when $p_k$ is closer to either 1 or 0). Therefore, a confidence score can be derived from $1-\mathcal{H}$, illustrating the model's confidence in this prediction.
Following~\citet{li2023mage}, we add a noise sampled from a random Gumbel distribution multiplied by the temperature $\tau$ to the confidence score.

At each sampling iteration, we select and unmask a proportion of masked positions with the highest confidence scores. To unmask each token, we obtain its binary codes by performing Bernoulli sampling from the distribution $\mathcal{B}(p_k)$.
The unmasking ratio follows a cosine schedule as used in~\citet{chang2022maskgit,li2023mage}.
This process operates over $N$ sampling iterations.
When the mask ratio drops to zero, the sampling progresses to the last iteration where all tokens are unmasked, marking the completion of the generation process.

\section{Experiment}
\subsection{Implementation details}
\textbf{Model configuration}\quad
We use the binary autoencoder (B-AE) introduced by~\citet{wang2023binary} as our binary tokenizer. The downsampling rate of the autoencoder is 16, projecting a 256$\times$256 image into a 16$\times$16 token sequence. We train four variants of the binary autoencoders designed with four different binary code dimensions, namely 16, 20, 24, and 32.
With the four binary tokenizers, we train our \modelname of three different sizes based on Llama~\citep{touvron2023llama1}, namely L (316M), XL (743M), and XXL (1.38B). 
For the binary transcoder, we follow~\citet{li2024autoregressive} to employ an MLP $g_\phi$ with an adaptive LayerNorm, with sizes of 20M, 56M, and 104M respectively. 
For clarity, we denote the \texttt{S}-sized variant with a \texttt{B}-dim tokenizer as \modelname-\texttt{S}-d\texttt{B}, \eg, \modelname-L-d16.

\textbf{Training details}\quad
We train all models for 400 epochs, with L-sized models using a batch size of 1024 and the others using a batch size of 512. 
Our L/XL-sized models are trained on 8 A800 GPUs, while XXL-sized models are trained on 32 A800 GPUs. 
We maintain consistent training settings across all compared models based on the model size.

\textbf{Sampling}\quad
Our model inherently supports classifier-free guidance (CFG)~\citep{ho2022classifier} through the Bernoulli diffusion process. Within our sampling process, four hyperparameters are involved: CFG scale, Gumbel temperature ($\tau$), the number of sampling iterations ($N$), and the number of Bernoulli denoising steps ($T$). We identify the optimal hyperparameter setting for all models. We set CFG to 2.5 for all quantitative evaluations, which has shown to be effective across all our models. 
We set $T$ to 100 as default for all models. See more details in~\Cref{appx:detail}.

\vspace{-5pt}
\subsection{Uniform performance}
\vspace{-5pt}
\textbf{Evaluation}\quad
We evaluate the uniformity of \modelname by concurrently comparing generative and discriminative performance. 
We evaluate \textbf{generation quality} on ImageNet-1K 256$\times$256 by reporting Frechet Inception Distance (FID) as the main metric, along with Inception Score (IS), sFID, Precision (Pre.), and Recall (Rec.) as auxiliary metrics. All metrics are obtained using 50K generated samples. 
We assess \textbf{representation capabilities} through linear-probe evaluation, reporting the top-1 and top-5 accuracies, abbreviated as ACC1 and ACC5, on ImageNet-1k 256$\times$256 validation split. 
We follow standard practice~\citep{he2022masked} by using a parameter-free BatchNorm~\citep{ioffe2015batch} layer and a linear classifier layer to classify the model features. 
We use the intermediate features from the 10-th layer for L-sized models, the 15-th layer for XL-sized models, and the 32-nd layer for XXL-sized models, as experiments on d16 models demonstrate these configurations yield the best performance. 
Additionally, we compare the inference speed, specifically the time taken to generate each image using one 4090 GPU with a batch size of 64.

\textbf{Comparison}\quad
Starting from the latest autoregressive generation baseline LlamaGen~\citep{sun2024autoregressive}, we comprehensively analyze two major components in this paper, namely (1) training objectives, specifically categorical loss (cat.) and binary loss (bin.), and (2) modeling types, including masking and autoregressive (AR) approaches. In total, we compare five models in~\Cref{tab:cross-compare}, training four models—S0, S1, S2, and S3—with different configurations, excluding LlamaGen. 
For LlamaGen, \rebut{we use the generative performance results under a \emph{fair} setting of 256$\times$256 image generation, as reported in their paper.} We conduct our own evaluation of linear-probe performance using their pretrained model. The inference time of all models is tested on the same machines by us.

\textbf{Observation}\quad
As shown in~\Cref{tab:cross-compare}, our model significantly outperforms other methods across all main evaluation metrics. In addition, we have the following observations.
\textbf{(1)} By comparing LlamaGen and S0, using binary autoencoder provides better generative performance and worse discriminative performance compared to VQGAN.
\textbf{(2)} For generation, AR modeling is better suited for categorical loss, while masked modeling is more appropriate for binary loss. 
\textbf{(3)} For discrimination, masked modeling drastically outperforms AR modeling for both losses, with binary loss further enhancing performance.
\textbf{(4)} Masked modeling achieves significantly faster inference speed compared to AR modeling due to its fewer sampling iterations, with the binary objective taking more time resulting from the diffusion process. This can be further accelerated by reducing sampling iterations and diffusion timesteps, as discussed in~\Cref{fig:iters}.
To conclude, \modelname, which employs masked modeling on binary latent codes, achieves the best \textit{uniform} performance on both generative and discriminative tasks, accompanied by an efficient inference runtime.

\begin{table}[t]
   \centering
   \caption{\textbf{Uniformity comparison.}
    We compare the generative and discriminative performance of our model against LlamaGen~\citep{sun2024autoregressive} and three other settings, varying by tokenizers, training objectives, and modeling types. We use KV cache to accelerate all AR models.} 
    \setlength{\tabcolsep}{7pt} 
    \scalebox{0.75}{
    \begin{tabular}{lccccccccccc}
        \toprule
         & & & & & \multicolumn{5}{c}{Generative} & \multicolumn{2}{c}{Discriminative} \\
        \cmidrule(lr){6-10}  \cmidrule(lr){11-12}
        Model & Tokenizer & Objective & Type & Time$\downarrow$ & FID$\downarrow$ & IS$\uparrow$ & sFID$\downarrow$ & Pre.$\uparrow$ & Rec.$\uparrow$ & ACC1 & ACC5 \\
        \midrule
        LlamaGen & VQGAN & Cat. & AR & 0.13 & 3.81 & 248.28 & 8.49 & 0.83 & 0.52 &  40.5 & 64.4 \\
        \midrule
        S0 & B-AE & Cat. & AR & 0.15 & 3.21 & 239.17 & \textbf{5.38} & 0.83 & \textbf{0.54} &  23.8 & 44.2 \\
        S1 & B-AE & Cat. & Mask & \textbf{0.10} & 3.85 & 261.81 & 6.10 & 0.85 & 0.47 &  61.1 & 83.2 \\
        \midrule
        S2 & B-AE & Bin. & AR & 1.04 & 7.50 & 164.31 & 6.56 & 0.85 & 0.41 &  45.2 & 69.3 \\
        S3 (Ours) & B-AE & Bin. & Mask & 0.69 & \textbf{3.17} & \textbf{262.14} & 5.59 & \textbf{0.86} & 0.50 & \textbf{64.3} & \textbf{85.4} \\
        \bottomrule
    \end{tabular}}
    \label{tab:cross-compare}
\end{table}

\vspace{-5pt}
\subsection{Model analysis}
\vspace{-5pt}
We analyze each component of our proposed method below.
All experiments are conducted on \modelname-L-d16 unless otherwise specified.

\textbf{Binary transcoder}\quad
We apply Bernoulli denoising process~\citep{wang2023binary} as our binary transcoder to generate probabilities of Bernoulli distributions, from which the binary codes are sampled. 
We experiment with two variants, namely (1) predicting the initial clean latent $z_0$, and (2) predicting the element-wise exclusive OR (XOR) value between the latent at the $t$-th timestep $z^t$ and $z^0$. We find empirically the latter performs better, and thus, we adopt this setting for all of our models.
Alternatively, a na\"ive approach involves using a direct binary cross-entropy (BCE) loss to train the model, replacing the Bernoulli denoising process. We compare these three variants in~\Cref{tab:binary-transcoder}. Our method outperforms the other two variants across all main metrics. We observe that using direct BCE generates very smooth images which harms the generative performance. XOR prediction yields better generative and discriminative performance compared to $z^0$ prediction.

\begin{table}[t]
\begin{minipage}{0.48\textwidth}
\centering
   \caption{\textbf{Binary transcoder comparison.}} 
    \scalebox{0.5}{
    \begin{tabular}{lccccccc}
        \toprule
        & \multicolumn{5}{c}{Generative} & \multicolumn{2}{c}{Discriminative} \\
        \cmidrule(lr){2-6}  \cmidrule(lr){7-8}
        Binary objective & FID$\downarrow$ & IS$\uparrow$ & sFID$\downarrow$ & Pre.$\uparrow$ & Rec.$\uparrow$ & ACC1 & ACC5 \\
        \midrule
        \emph{w/o Bernoulli denoising} \\
        \quad Direct BCE & 5.84 & 212.34 & 9.89 & 0.78 & \textbf{0.52} & 63.3 & 84.8 \\
        \emph{w/ Bernoulli denoising} \\
        \quad Predict $z^0$ & 4.39 & \textbf{274.26} & 9.07 & \textbf{0.87} & 0.44 & 62.0 & 83.9 \\
        \quad Predict $z^{t} \oplus z^0$ (Ours) & \textbf{3.17} & 262.14 & \textbf{5.59} & 0.86 & 0.50 & \textbf{64.3} & \textbf{85.4} \\
        \bottomrule
    \end{tabular}}
    \label{tab:binary-transcoder} 
\end{minipage}
\ \ 
\begin{minipage}{0.48\textwidth}
   \centering
    \caption{\textbf{Sampling order comparison.} We include the autoregressive variant for reference.} 
    \scalebox{0.57}{
    \begin{tabular}{lccccccc}
        \toprule
        Type & Order & Time$\downarrow$ & FID$\downarrow$ & IS$\uparrow$ & sFID$\downarrow$ & Pre.$\uparrow$ & Rec.$\uparrow$ \\
        \midrule
        \textcolor{gray}{AR} & \textcolor{gray}{Raster} & \textcolor{gray}{1.04} & \textcolor{gray}{7.50} & \textcolor{gray}{164.31} & \textcolor{gray}{6.56} & \textcolor{gray}{0.85} & \textcolor{gray}{0.41} \\
        Mask & Raster & 8.81 & 4.51 & 191.10 & 6.49 & 0.80 & 0.54 \\
        Mask & Rand. & \textbf{0.69} & 7.12 & 174.11 & 11.85 & 0.76 & \textbf{0.55} \\
        Mask & Ours & \textbf{0.69} & \textbf{3.17} & \textbf{262.14} & \textbf{5.59} & \textbf{0.86} & 0.50 \\  
        \bottomrule
    \end{tabular}}
    \label{tab:sampling-order} 
    \end{minipage}
\end{table}

\textbf{Sampling strategy}\quad
In this paper, we propose a simple entropy-ordered sampling strategy tailored for the masked training paradigm. We compare our method with two alternative sampling orders, namely (1) a raster-scan order similar to the autoregressive approach, and (2) a random order. Like our strategy, both compared methods are applied to the same trained model. The comparison results of the generative evaluation are reported in~\Cref{tab:sampling-order}. The results indicate that the proposed sampling strategy is the best fit for our model's generative purposes. 

\begin{figure}[t]
    \centering
    \includegraphics[width=1.0\linewidth]{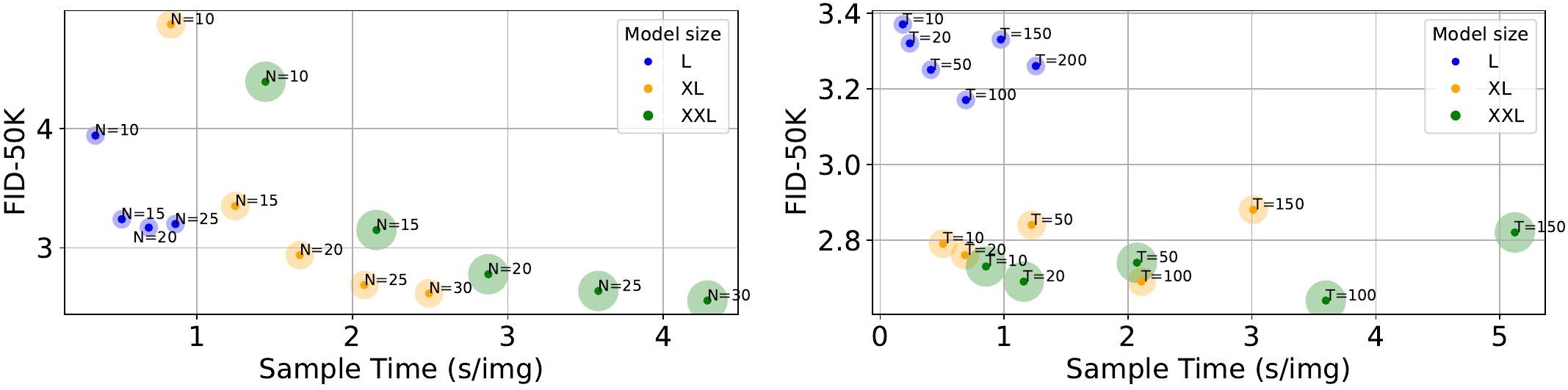}
    \caption{\textbf{Relationships between FID-50K and sample time across varying inference hyperparameters.} We compare different numbers of sampling iterations $N$ (left) and varying diffusion timesteps $T$ (right) for three model sizes. All other hyperparameters are kept at their default settings.}
    \label{fig:iters}
    \vspace{-5pt}
\end{figure}

\textbf{Inference hyperparameters}\quad
We evaluate the impact of two hyperparameters specific to our model on its performance. 
\textbf{(1)} We first present the FID results and sample time for different numbers of sampling iterations $N$ on the left side of~\Cref{fig:iters}. 
We observe that larger models generally achieve lower FID values, although they also increase sample time. In addition, more sampling iterations do not guarantee better performance, as different-sized models have varying optimal sampling iterations. For example, the L-sized model achieves its best performance with 20 iterations, rather than with larger numbers. 
\textbf{(2)} On the right side of~\Cref{fig:iters}, we present the results for different numbers of diffusion timesteps $T$. The results indicate that diffusion timesteps have a marginal impact on generative performance, suggesting that our model can achieve comparable generation quality with fewer diffusion timesteps. This can significantly accelerate generation speed, especially for larger models. For example, with 10 diffusion timesteps, the XXL-sized model can achieve an FID of 2.73 at a speed of 0.85s per image.

\begin{figure}[t]
    \centering
    \includegraphics[width=1.0\linewidth]{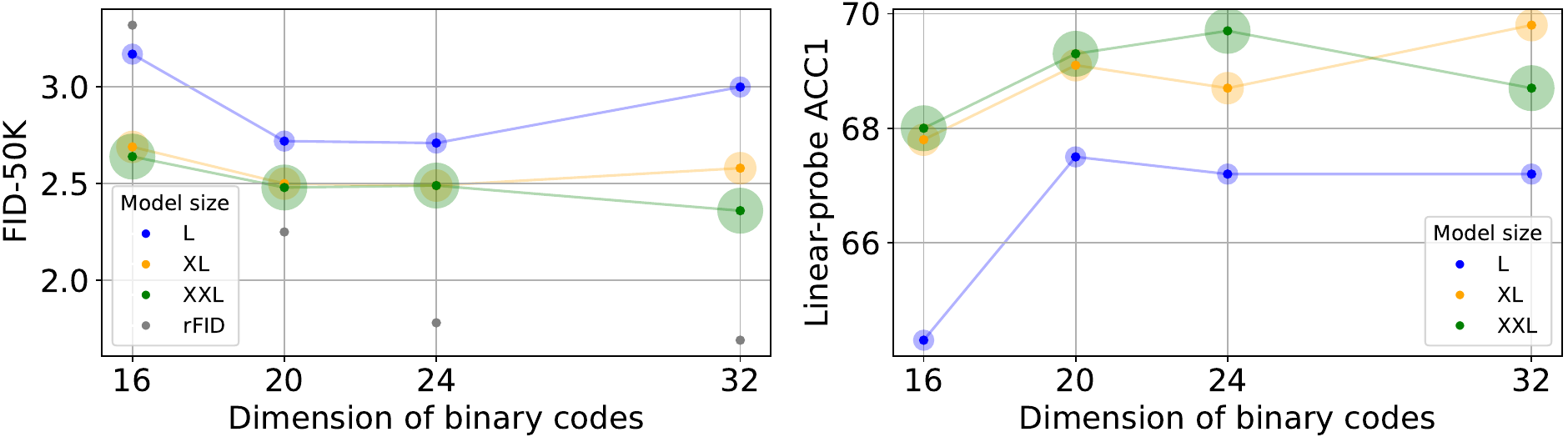}
    \caption{\textbf{Evaluation of generative and discriminative performance across different model sizes.} We report results for all tested tokenizers across four different dimensions of binary codes. We include the reconstruction FID (rFID) for each binary tokenizer for reference (grey points).}
    \label{fig:scalibility}
    \vspace{-5pt}
\end{figure}

\textbf{Model size and code dimension}\quad
We validate that our model is scalable by testing the performance of different-sized models using tokenizers with various code dimensions. Note that the dimension of the binary codes only alters the number of parameters in the input and output linear projections, resulting in minimal effects on the overall model size. 
The evaluation results of both generative and discriminative performance are shown in~\Cref{fig:scalibility}. 
Our model generally performs better with larger sizes across all code dimensions, as indicated by both generative and discriminative metrics. 

Besides, we have the following observations  from~\Cref{fig:scalibility}.
\textbf{(1)} When the model size is small, it becomes challenging to model large-dimensional codes, such as a dimension of 32 for the L-sized model, especially for generative purpose. 
\textbf{(2)} In contrast, as the model size increases, the improvement for smaller-dimensional codes is relatively modest, indicating that these codes are easier to model and can be effectively handled by smaller-sized models.
\textbf{(3)} An exception arises in the linear-probe evaluation of models with 32-dimensional codes, where our XL-sized model outperforms the XXL-sized model. We hypothesize that this may be due to the optimal transformer layer for feature representation identified in the 16-dimensional model, which might not be the best choice for 32-dimensional models of the XXL size.

\begin{figure}[t]
\begin{minipage}{0.53\textwidth}
    \centering
    \includegraphics[width=1.0\linewidth]{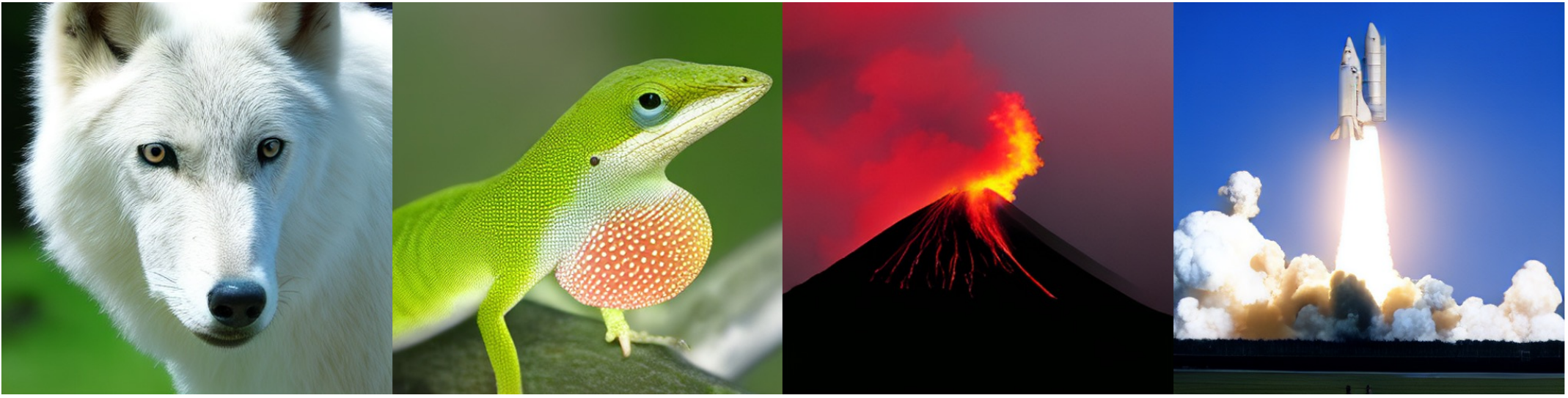}
    \caption{\textbf{Generated 512$\times$512 samples.}}
    \label{fig:res512}
\end{minipage}
\hfill
\begin{minipage}{0.45\textwidth}
    \setlength{\tabcolsep}{10pt} 
    \centering
    \captionof{table}{\textbf{Linear-probe evaluation} of conditional and unconditional counterparts.}
    \scalebox{0.8}{
    \begin{tabular}{lccccc}
        \toprule
        Training & ACC1 & ACC5 \\
        \midrule
        Cond. &  67.5 & 87.5 \\
        Uncond. &  \textbf{68.3} & \textbf{88.4} \\
        \bottomrule
    \end{tabular}}
    \label{tab:uncond}  
\end{minipage}
\vspace{-5pt}
\end{figure}

\textbf{Unconditional training}\quad
Our model is a class-conditional generative model. 
Intuitively, conditional generative training adds condition guidance that is absent in downstream discriminative tasks, which can diminish the representation capabilities of the model. 
We validate this conjecture by comparing the linear-probe performance of our model with that of its unconditional counterpart.
We train the unconditional model by replacing the class conditional tokens with a single unconditional token, and keep the inference process unchanged. We evaluate BiGR-L-d20 alongside its unconditional counterpart, and report the results in~\Cref{tab:uncond}. 
The unconditional counterpart demonstrates better representation capabilities than our conditional model, indicating that discriminative tasks are more challenging for conditional generative models.

\textbf{Resolution of 512$\times$512}\quad
Using a binary autoencoder that projects a 512$\times$512 image into 32$\times$32 binary latent codes, we enable our model to generate 512$\times$512 images by increasing the input sequence length to 1024. 
We train such a binary autoencoder with a code dimension of 32 and train our model to accommodate this sequence length. 
We showcase the generated samples in~\Cref{fig:res512}, with additional samples available in~\Cref{appx:more}.

\vspace{-10pt}
\subsection{System-level comparisons}
\vspace{-5pt}
We re-emphasize that the goal of this work is to propose a uniform conditional generative model that can produce high-quality generations while maintaining strong representation capabilities. 
Therefore, surpassing state-of-the-art models across all metrics is not within the scope of this research. 
We provide a more comprehensive comparison in~\Cref{appx:compare-more}.

\textbf{Conditional image generation}\quad
We present a comparison of the generative performance of our model with leading generative systems in~\Cref{tab:compare-gen}. 
\rebut{The recent MAR~\citep{li2024autoregressive} achieves the lowest FID, and SiT~\citep{ma2024sit} sets the state-of-the-art among diffusion-based models.}
Our model maintains top-tier generative quality among the first echelon of approaches. 
Besides, \modelname significantly outperforms LlamaGen.

\begin{table}[t]
\begin{minipage}{0.49\textwidth}
\centering
\caption{\textbf{Generative performance comparison} on 256$\times$256 ImageNet-1K benchmark.}
\label{tab:compare-gen}
\scalebox{0.59}{
\begin{tabular}{ll|c|cccc}
\toprule
\textbf{Type} & \textbf{Model} & \textbf{\#Params.} & \textbf{FID$\downarrow$} & \textbf{IS$\uparrow$} \\
\midrule
\multirow{4}{*}{\textbf{Diff.}} 
% & ADM (Dhariwal and Nichol 2021a) & 554M & 10.94 & 101.0 & 0.69 & 0.63 \\
 & DiT-L/2~\citep{peebles2023scalable} & 458M & 5.02 & 167.2 \\
 & DiT-XL/2 & 675M & 2.27 & 278.2 \\
 & \rebut{SiT-XL/2 (ODE)~\citep{ma2024sit}} & \rebut{675M} & \rebut{2.15} & \rebut{258.1} \\
 & \rebut{SiT-XL/2 (SDE)} & \rebut{675M} & \rebut{2.06} & \rebut{277.5} \\
\midrule
\multirow{1}{*}{\textbf{Mask}} & MaskGIT \citep{chang2022maskgit} & 227M & 6.18 & 182.1 \\
\midrule
\multirow{4}{*}{\textbf{AR}} & VQGAN \citep{esser2021taming} & 227M & 18.65 & 80.4 \\
 & VQGAN & 1.4B  & 15.78 & 74.3  \\
 & ViT-VQGAN \citep{yu2022vector} & 1.7B & 4.17 & 175.1 \\
 & RQTran. \citep{lee2022autoregressive} & 3.8B & 7.55 & 134.0 \\
\midrule
\multirow{4}{*}{\textbf{VAR}} & VAR-d16 \citep{tian2024visual} & 310M & 3.30 & 274.4 \\
 & VAR-d20 & 600M & 2.57 & 302.6 \\
 & VAR-d24 & 1.0B & 2.09 & 312.9 \\
 & VAR-d30 & 2.0B & 1.92 & 323.1 \\
\midrule
 \multirow{3}{*}{\textbf{MAR}} & MAR-B \citep{li2024autoregressive} & 208M & 2.31 & 281.7 \\
 & MAR-L & 479M & 1.78 & 296.0 \\
 & MAR-H & 943M & 1.55 & 303.7 \\
\midrule
\multirow{5}{*}{\textbf{AR}} & LlamaGen-B \citep{sun2024autoregressive} & 111M & 5.46 & 193.6 \\
 & LlamaGen-L & 343M & 3.81 & 248.3 \\
 & LlamaGen-XL & 775M & 3.39 & 227.1 \\
 & LlamaGen-XXL & 1.4B & 3.09 & 253.6 \\
 & LlamaGen-3B & 3.1B & 3.05 & 222.3 \\
 \midrule
\multirow{3}{*}{\textbf{Ours}} & \modelname-L-d24 & 336M & 2.71 & 275.7 \\
& \modelname-XL-d24 & 799M & 2.49 & 278.8 \\
& \modelname-XXL-d32 & 1.5B & 2.36 & 277.2 \\
\bottomrule
\end{tabular}}
\end{minipage}
\ 
\begin{minipage}{0.49\textwidth}
\centering
\caption{\textbf{Linear-probe top-1 accuracy} on ImageNet-1K. $^\dag$: our evaluation results.}
\label{tab:linprobe}
\scalebox{0.61}{
\begin{tabular}{ll|cc|c}
\toprule
\textbf{Type} & \textbf{Method} & \textbf{\#Tokens} & \textbf{Params} & \textbf{ACC1$\uparrow$} \\
\midrule
\multirow{8}{*}{\textbf{Con.}}
& MoCo~\citep{he2020moco} & - & 375M & 68.6 \\
& SimCLR~\citep{chen2020simclr} & - & 375M & 76.5 \\
& SwAV~\citep{caron2020swav} & - & 93M & 75.3 \\
& DINO~\citep{caron2021dino} & - & 85M & 75.3 \\
& BYOL~\citep{grill2020byol} & - & 375M & 78.6 \\
& CAE~\citep{chen2024context} & - & 304M & 78.1 \\
& CMAE~\citep{huang2023contrastive} & - & 86M & 73.9 \\
\midrule
\multirow{4}{*}{\textbf{MIM}}
& iBOT~\citep{zhou2022ibot} & - & 304M & 81.0 \\
& BEiT~\citep{bao2022beit} & 16$\times$16 & 307M & 73.5 \\
& MAE~\citep{he2022masked} & 14$\times$14 & 304M & 75.8 \\
& MAGE~\citep{li2023mage} & 16$\times$16 & 328M & 78.9 \\
\midrule
\multirow{9}{*}{\textbf{Gen.}}
& BigBiGAN~\citep{brock2018large} & - & 344M & 61.3 \\
& iGPT-L~\citep{chen2020generative} & 32$\times$32 & 1.4B & 60.3 \\
& iGPT-L & 48$\times$48 & 1.4B & 65.2 \\
& ViT-VQGAN-B~\citep{yu2022vector} & 32$\times$32 & 650M & 65.1 \\
& ViT-VQGAN-L & 32$\times$32 & 1.7B & 73.2 \\
& RCG~\citep{li2023self} & 16$\times$16 & 304M & 77.6 \\
& \emph{l}-DAE~\citep{chen2024deconstructing} & - & 304M & 75.0 \\
\midrule
\multirow{6}{*}{\shortstack{\textbf{Cond.} \\ \textbf{gen.}}}
& LlamaGen-L$^\dag$~\citep{sun2024autoregressive} & 16$\times$16 & 343M & 40.5 \\
& MAR-B$^\dag$~\citep{li2024autoregressive} & 16$\times$16 & 208M & 57.9 \\
& MAR-L$^\dag$ & 16$\times$16 & 479M & 59.1 \\
& MAR-H$^\dag$ & 16$\times$16 & 943M & 60.0 \\
& \modelname-L-d20 (Ours) & 16$\times$16 & 336M & 67.5 \\
& \modelname-XL-d32 (Ours) & 16$\times$16 & 799M & 69.8 \\
% & \modelname-XXL-d20 (Ours) & 16$\times$16 & \\
\bottomrule
\end{tabular}}
\end{minipage}
\end{table}

\textbf{Visual representation}\quad
We compare the linear-probe results of our model and the previous methods specifically designed for discriminative tasks. The results are shown in~\Cref{tab:linprobe}.
We categorize the compared models into several types: contrastive (Con.), masked image modeling (MIM), generative (Gen.), and conditional generative (Cond. gen.).
This classification is not entirely precise, as some models may use multiple losses for training, like MAGE~\citep{li2023mage}. Our model is fairly compared to the conditional generative models, which solely rely on plain reconstruction loss without discriminative designs, such as specialized losses, augmentations, or additional data. For LlamaGen~\citep{sun2024autoregressive} that has the same model architecture as ours, we use the feature from the same layer for linear layer training. For MAR~\citep{li2024autoregressive}, since their structure largely resembles MAE~\citep{he2022masked}, we follow MAE's approach and train the linear layer on top of the encoder outputs. \modelname significantly outperforms the other conditional generative models. 

\vspace{-3pt}
\subsection{Zero-shot generalized applications}
\begin{figure}[t]
    \centering
    \includegraphics[width=1.0\linewidth]{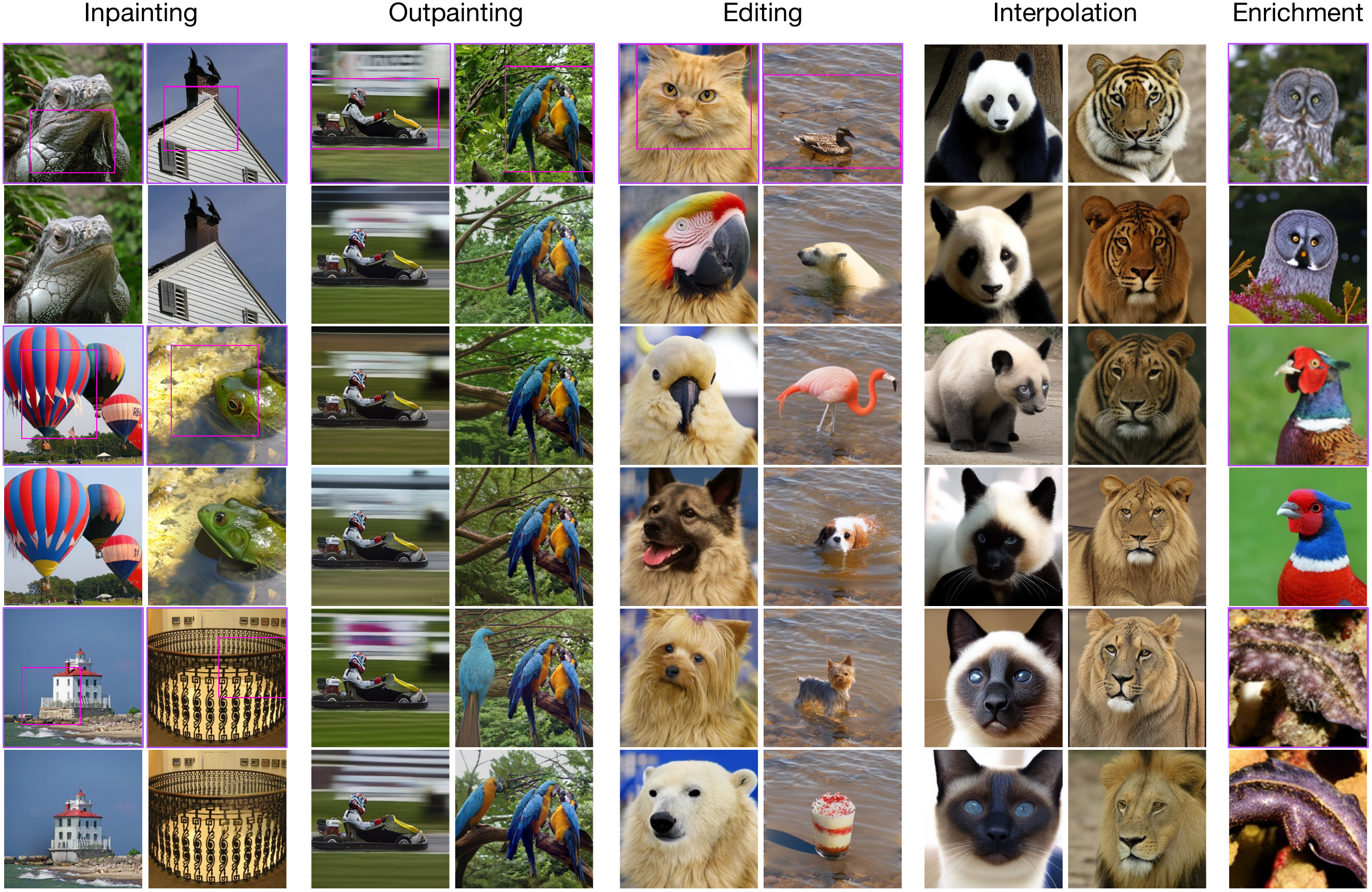}
    \caption{\textbf{Zero-shot generalization.} We present samples of inpainting, outpainting, editing, interpolation, and enrichment. The original image is marked with a purple border, with a pink box highlighting the masked region. Images without the purple borders are generated by our model.}
    \label{fig:apps}
\end{figure}

The nature of the masked modeling mechanism allows the use of \modelname in a wide range of applications in a zero-shot manner, without the need for task-specific structural changes or parameter fine-tuning. We present the results of \modelname applied across various tasks in~\Cref{fig:apps}.

\textbf{Inpainting \& Outpainting}\quad
Given an image with a mask, we use the unmasked regions to initialize the model inputs, enabling it to generate the remaining masked tokens. This generation process is guided by an unconditional token, which ensures that no additional information is introduced, allowing the model to focus solely on the existing image information. This approach enables high-quality and diverse inpainting and outpainting.

\textbf{Class-conditional editing}\quad
Unlike inpainting and outpainting, class-conditional editing is guided by a specific class condition, allowing the model to edit the masked region with a designated class object. Other operations remain consistent with inpainting and outpainting.

\textbf{Class interpolation}\quad
We interpolate between two class conditions by calculating a weighted sum in the embedding space. We then use the resulting interpolated embedding to guide the generation process. This interpolation process demonstrates that our model can generalize visual characteristics across different classes rather than merely memorizing each class.

\textbf{Image enrichment}\quad
Our model can also enrich visual details in a low-resolution image, a process we call image enrichment.
Specifically, we first upsample a 128$\times$128 image to a resolution of 256$\times$256 and encode it into a sequence of 16$\times$16 tokens. This approach leverages the model's generative capabilities to enrich images from low-resolution inputs.
\vspace{-5pt}
\section{Conclusion}
We introduce \modelname as the first conditional generative model that unifies generative and discriminative tasks within the same framework. Through extensive experiments, we highlight its uniformity, efficiency, flexibility, and scalability. Our results demonstrate that \modelname achieves decent performance in both generation quality and linear separability. Additionally, we showcase its application in various zero-shot generalized tasks. We believe \modelname has the potential to be adapted for a broader range of applications in the future.

\textbf{Limitations}\quad
(1) Our sampling strategy involves numerous hyperparameters to tune, resulting in a substantial search space; thus, the reported models may not represent the optimal settings.
(2) The model's sequence length is fixed during training, making it inflexible to accommodate inputs of varying lengths. Consequently, generating higher-resolution images requires re-training the model.

\paragraph{Acknowledgement} This work is partially supported by the Hong Kong Research Grants Council - General Research Fund (Grant No.: 17211024).

% \textbf{Future work}\quad
% Given the promising performance of \modelname, it can be applied to more general text-to-image generation and multi-modal tasks. Leveraging \modelname's representation capabilities, the text-to-image or multi-modal model can achieve stronger discriminative features, benefiting a wide range of downstream tasks such as recognition, retrieval, alignment, and \etc.
% We believe our model has the potential to be adapted for a broader range of applications in the future.



\bibliography{iclr2025_conference}

\begin{thebibliography}{59}
\providecommand{\natexlab}[1]{#1}
\providecommand{\url}[1]{\texttt{#1}}
\expandafter\ifx\csname urlstyle\endcsname\relax
  \providecommand{\doi}[1]{doi: #1}\else
  \providecommand{\doi}{doi: \begingroup \urlstyle{rm}\Url}\fi

\bibitem[Balestriero \& LeCun(2024)Balestriero and LeCun]{balestriero2024how}
Randall Balestriero and Yann LeCun.
\newblock How learning by reconstruction produces uninformative features for perception.
\newblock In \emph{ICML}, 2024.

\bibitem[Bao et~al.(2022)Bao, Dong, Piao, and Wei]{bao2022beit}
Hangbo Bao, Li~Dong, Songhao Piao, and Furu Wei.
\newblock {BE}it: {BERT} pre-training of image transformers.
\newblock In \emph{ICLR}, 2022.

\bibitem[Betker et~al.(2023)Betker, Goh, Jing, Brooks, Wang, Li, Ouyang, Zhuang, Lee, Guo, et~al.]{betker2023dalle3}
James Betker, Gabriel Goh, Li~Jing, Tim Brooks, Jianfeng Wang, Linjie Li, Long Ouyang, Juntang Zhuang, Joyce Lee, Yufei Guo, et~al.
\newblock Improving image generation with better captions.
\newblock \emph{Computer Science. https://cdn. openai. com/papers/dall-e-3. pdf}, 2023.

\bibitem[Brock(2018)]{brock2018large}
Andrew Brock.
\newblock Large scale gan training for high fidelity natural image synthesis.
\newblock \emph{arXiv preprint arXiv:1809.11096}, 2018.

\bibitem[Brown et~al.(2020)Brown, Mann, Ryder, Subbiah, Kaplan, Dhariwal, Neelakantan, Shyam, Sastry, Askell, Agarwal, Herbert-Voss, Krueger, Henighan, Child, Ramesh, Ziegler, Wu, Winter, Hesse, Chen, Sigler, Litwin, Gray, Chess, Clark, Berner, McCandlish, Radford, Sutskever, and Amodei]{gpt3}
Tom Brown, Benjamin Mann, Nick Ryder, Melanie Subbiah, Jared~D Kaplan, Prafulla Dhariwal, Arvind Neelakantan, Pranav Shyam, Girish Sastry, Amanda Askell, Sandhini Agarwal, Ariel Herbert-Voss, Gretchen Krueger, Tom Henighan, Rewon Child, Aditya Ramesh, Daniel Ziegler, Jeffrey Wu, Clemens Winter, Chris Hesse, Mark Chen, Eric Sigler, Mateusz Litwin, Scott Gray, Benjamin Chess, Jack Clark, Christopher Berner, Sam McCandlish, Alec Radford, Ilya Sutskever, and Dario Amodei.
\newblock Language models are few-shot learners.
\newblock In \emph{NeurIPS}, 2020.

\bibitem[Cakir et~al.(2019)Cakir, He, Bargal, and Sclaroff]{cakir2019hashing}
Fatih Cakir, Kun He, Sarah~Adel Bargal, and Stan Sclaroff.
\newblock Hashing with mutual information.
\newblock \emph{IEEE TPAMI}, 2019.

\bibitem[Caron et~al.(2020)Caron, Misra, Mairal, Goyal, Bojanowski, and Joulin]{caron2020swav}
Mathilde Caron, Ishan Misra, Julien Mairal, Priya Goyal, Piotr Bojanowski, and Armand Joulin.
\newblock Unsupervised learning of visual features by contrasting cluster assignments.
\newblock In \emph{NeurIPS}, 2020.

\bibitem[Caron et~al.(2021)Caron, Touvron, Misra, J\'egou, Mairal, Bojanowski, and Joulin]{caron2021dino}
Mathilde Caron, Hugo Touvron, Ishan Misra, Herv\'e J\'egou, Julien Mairal, Piotr Bojanowski, and Armand Joulin.
\newblock Emerging properties in self-supervised vision transformers.
\newblock In \emph{ICCV}, 2021.

\bibitem[Chang et~al.(2022)Chang, Zhang, Jiang, Liu, and Freeman]{chang2022maskgit}
Huiwen Chang, Han Zhang, Lu~Jiang, Ce~Liu, and William~T Freeman.
\newblock Maskgit: Masked generative image transformer.
\newblock In \emph{CVPR}, 2022.

\bibitem[Chang et~al.(2023)Chang, Zhang, Barber, Maschinot, Lezama, Jiang, Yang, Murphy, Freeman, Rubinstein, Li, and Krishnan]{chang2023muse}
Huiwen Chang, Han Zhang, Jarred Barber, Aaron Maschinot, Jose Lezama, Lu~Jiang, Ming-Hsuan Yang, Kevin~Patrick Murphy, William~T. Freeman, Michael Rubinstein, Yuanzhen Li, and Dilip Krishnan.
\newblock Muse: Text-to-image generation via masked generative transformers.
\newblock In \emph{ICML}, 2023.

\bibitem[Chen et~al.(2024{\natexlab{a}})Chen, Jincheng, Chongjian, Yao, Xie, Wang, Kwok, Luo, Lu, and Li]{chenpixart}
Junsong Chen, YU~Jincheng, GE~Chongjian, Lewei Yao, Enze Xie, Zhongdao Wang, James Kwok, Ping Luo, Huchuan Lu, and Zhenguo Li.
\newblock Pixart-$\alpha$: Fast training of diffusion transformer for photorealistic text-to-image synthesis.
\newblock In \emph{ICLR}, 2024{\natexlab{a}}.

\bibitem[Chen et~al.(2024{\natexlab{b}})Chen, YU, GE, Yao, Xie, Wang, Kwok, Luo, Lu, and Li]{chen2024pixartalpha}
Junsong Chen, Jincheng YU, Chongjian GE, Lewei Yao, Enze Xie, Zhongdao Wang, James Kwok, Ping Luo, Huchuan Lu, and Zhenguo Li.
\newblock Pixart-\${\textbackslash}alpha\$: Fast training of diffusion transformer for photorealistic text-to-image synthesis.
\newblock In \emph{ICLR}, 2024{\natexlab{b}}.

\bibitem[Chen et~al.(2020{\natexlab{a}})Chen, Radford, Child, Wu, Jun, Luan, and Sutskever]{chen2020generative}
Mark Chen, Alec Radford, Rewon Child, Jeffrey Wu, Heewoo Jun, David Luan, and Ilya Sutskever.
\newblock Generative pretraining from pixels.
\newblock In \emph{ICML}, 2020{\natexlab{a}}.

\bibitem[Chen et~al.(2020{\natexlab{b}})Chen, Kornblith, Norouzi, and Hinton]{chen2020simclr}
Ting Chen, Simon Kornblith, Mohammad Norouzi, and Geoffrey Hinton.
\newblock A simple framework for contrastive learning of visual representations.
\newblock In \emph{ICML}, 2020{\natexlab{b}}.

\bibitem[Chen et~al.(2024{\natexlab{c}})Chen, Ding, Wang, Xin, Mo, Wang, Han, Luo, Zeng, and Wang]{chen2024context}
Xiaokang Chen, Mingyu Ding, Xiaodi Wang, Ying Xin, Shentong Mo, Yunhao Wang, Shumin Han, Ping Luo, Gang Zeng, and Jingdong Wang.
\newblock Context autoencoder for self-supervised representation learning.
\newblock \emph{IJCV}, 2024{\natexlab{c}}.

\bibitem[Chen et~al.(2024{\natexlab{d}})Chen, Liu, Xie, and He]{chen2024deconstructing}
Xinlei Chen, Zhuang Liu, Saining Xie, and Kaiming He.
\newblock Deconstructing denoising diffusion models for self-supervised learning.
\newblock \emph{arXiv preprint arXiv:2401.14404}, 2024{\natexlab{d}}.

\bibitem[Donahue \& Simonyan(2019)Donahue and Simonyan]{donahue2019large}
Jeff Donahue and Karen Simonyan.
\newblock Large scale adversarial representation learning.
\newblock In \emph{NeurIPS}, 2019.

\bibitem[Donahue et~al.(2017)Donahue, Kr{\"a}henb{\"u}hl, and Darrell]{donahue2017adversarial}
Jeff Donahue, Philipp Kr{\"a}henb{\"u}hl, and Trevor Darrell.
\newblock Adversarial feature learning.
\newblock In \emph{ICLR}, 2017.

\bibitem[Dubey et~al.(2024)Dubey, Jauhri, Pandey, Kadian, Al-Dahle, Letman, Mathur, Schelten, Yang, Fan, et~al.]{dubey2024llama3}
Abhimanyu Dubey, Abhinav Jauhri, Abhinav Pandey, Abhishek Kadian, Ahmad Al-Dahle, Aiesha Letman, Akhil Mathur, Alan Schelten, Amy Yang, Angela Fan, et~al.
\newblock The llama 3 herd of models.
\newblock \emph{arXiv preprint arXiv:2407.21783}, 2024.

\bibitem[Esser et~al.(2021)Esser, Rombach, and Ommer]{esser2021taming}
Patrick Esser, Robin Rombach, and Bjorn Ommer.
\newblock Taming transformers for high-resolution image synthesis.
\newblock In \emph{CVPR}, 2021.

\bibitem[Grill et~al.(2020)Grill, Strub, Altch{\'e}, Tallec, Richemond, Buchatskaya, Doersch, Avila~Pires, Guo, Gheshlaghi~Azar, et~al.]{grill2020byol}
Jean-Bastien Grill, Florian Strub, Florent Altch{\'e}, Corentin Tallec, Pierre Richemond, Elena Buchatskaya, Carl Doersch, Bernardo Avila~Pires, Zhaohan Guo, Mohammad Gheshlaghi~Azar, et~al.
\newblock Bootstrap your own latent-a new approach to self-supervised learning.
\newblock In \emph{NeurIPS}, 2020.

\bibitem[He et~al.(2020)He, Fan, Wu, Xie, and Girshick]{he2020moco}
Kaiming He, Haoqi Fan, Yuxin Wu, Saining Xie, and Ross Girshick.
\newblock Momentum contrast for unsupervised visual representation learning.
\newblock In \emph{CVPR}, 2020.

\bibitem[He et~al.(2022)He, Chen, Xie, Li, Doll{\'a}r, and Girshick]{he2022masked}
Kaiming He, Xinlei Chen, Saining Xie, Yanghao Li, Piotr Doll{\'a}r, and Ross Girshick.
\newblock Masked autoencoders are scalable vision learners.
\newblock In \emph{CVPR}, 2022.

\bibitem[Henaff(2020)]{henaff2020data}
Olivier Henaff.
\newblock Data-efficient image recognition with contrastive predictive coding.
\newblock In \emph{ICML}, 2020.

\bibitem[Ho \& Salimans(2022)Ho and Salimans]{ho2022classifier}
Jonathan Ho and Tim Salimans.
\newblock Classifier-free diffusion guidance.
\newblock \emph{arXiv preprint arXiv:2207.12598}, 2022.

\bibitem[Ho et~al.(2020)Ho, Jain, and Abbeel]{ho2020denoising}
Jonathan Ho, Ajay Jain, and Pieter Abbeel.
\newblock Denoising diffusion probabilistic models.
\newblock In \emph{NeurIPS}, 2020.

\bibitem[Huang et~al.(2023)Huang, Jin, Lu, Hou, Cheng, Fu, Shen, and Feng]{huang2023contrastive}
Zhicheng Huang, Xiaojie Jin, Chengze Lu, Qibin Hou, Ming-Ming Cheng, Dongmei Fu, Xiaohui Shen, and Jiashi Feng.
\newblock Contrastive masked autoencoders are stronger vision learners.
\newblock \emph{IEEE TPAMI}, 2023.

\bibitem[Ioffe(2015)]{ioffe2015batch}
Sergey Ioffe.
\newblock Batch normalization: Accelerating deep network training by reducing internal covariate shift.
\newblock \emph{arXiv preprint arXiv:1502.03167}, 2015.

\bibitem[Jiang \& Li(2018)Jiang and Li]{jiang2018asymmetric}
Qing-Yuan Jiang and Wu-Jun Li.
\newblock Asymmetric deep supervised hashing.
\newblock In \emph{AAAI}, 2018.

\bibitem[Kang et~al.(2023)Kang, Zhu, Zhang, Park, Shechtman, Paris, and Park]{kang2023scaling}
Minguk Kang, Jun-Yan Zhu, Richard Zhang, Jaesik Park, Eli Shechtman, Sylvain Paris, and Taesung Park.
\newblock Scaling up gans for text-to-image synthesis.
\newblock In \emph{CVPR}, 2023.

\bibitem[Lee et~al.(2022)Lee, Kim, Kim, Cho, and Han]{lee2022autoregressive}
Doyup Lee, Chiheon Kim, Saehoon Kim, Minsu Cho, and Wook-Shin Han.
\newblock Autoregressive image generation using residual quantization.
\newblock In \emph{CVPR}, 2022.

\bibitem[Li et~al.(2023{\natexlab{a}})Li, Chang, Mishra, Zhang, Katabi, and Krishnan]{li2023mage}
Tianhong Li, Huiwen Chang, Shlok Mishra, Han Zhang, Dina Katabi, and Dilip Krishnan.
\newblock Mage: Masked generative encoder to unify representation learning and image synthesis.
\newblock In \emph{CVPR}, 2023{\natexlab{a}}.

\bibitem[Li et~al.(2023{\natexlab{b}})Li, Katabi, and He]{li2023self}
Tianhong Li, Dina Katabi, and Kaiming He.
\newblock Self-conditioned image generation via generating representations.
\newblock \emph{arXiv preprint arXiv:2312.03701}, 2023{\natexlab{b}}.

\bibitem[Li et~al.(2024)Li, Tian, Li, Deng, and He]{li2024autoregressive}
Tianhong Li, Yonglong Tian, He~Li, Mingyang Deng, and Kaiming He.
\newblock Autoregressive image generation without vector quantization.
\newblock \emph{arXiv preprint arXiv:2406.11838}, 2024.

\bibitem[Ma et~al.(2024)Ma, Goldstein, Albergo, Boffi, Vanden-Eijnden, and Xie]{ma2024sit}
Nanye Ma, Mark Goldstein, Michael~S Albergo, Nicholas~M Boffi, Eric Vanden-Eijnden, and Saining Xie.
\newblock Si{T}: Exploring flow and diffusion-based generative models with scalable interpolant transformers.
\newblock In \emph{ECCV}, 2024.

\bibitem[Peebles \& Xie(2023)Peebles and Xie]{peebles2023scalable}
William Peebles and Saining Xie.
\newblock Scalable diffusion models with transformers.
\newblock In \emph{ICCV}, 2023.

\bibitem[Raffel et~al.(2020)Raffel, Shazeer, Roberts, Lee, Narang, Matena, Zhou, Li, and Liu]{raffel2020exploring}
Colin Raffel, Noam Shazeer, Adam Roberts, Katherine Lee, Sharan Narang, Michael Matena, Yanqi Zhou, Wei Li, and Peter~J Liu.
\newblock Exploring the limits of transfer learning with a unified text-to-text transformer.
\newblock \emph{JMLR}, 2020.

\bibitem[Rombach et~al.(2022)Rombach, Blattmann, Lorenz, Esser, and Ommer]{rombach2022high}
Robin Rombach, Andreas Blattmann, Dominik Lorenz, Patrick Esser, and Bj{\"o}rn Ommer.
\newblock High-resolution image synthesis with latent diffusion models.
\newblock In \emph{CVPR}, 2022.

\bibitem[Saharia et~al.(2022)Saharia, Chan, Saxena, Li, Whang, Denton, Ghasemipour, Gontijo~Lopes, Karagol~Ayan, Salimans, et~al.]{saharia2022imagen}
Chitwan Saharia, William Chan, Saurabh Saxena, Lala Li, Jay Whang, Emily~L Denton, Kamyar Ghasemipour, Raphael Gontijo~Lopes, Burcu Karagol~Ayan, Tim Salimans, et~al.
\newblock Photorealistic text-to-image diffusion models with deep language understanding.
\newblock In \emph{NeurIPS}, 2022.

\bibitem[Sauer et~al.(2022)Sauer, Schwarz, and Geiger]{sauer2022stylegan}
Axel Sauer, Katja Schwarz, and Andreas Geiger.
\newblock Stylegan-xl: Scaling stylegan to large diverse datasets.
\newblock In \emph{ACM SIGGRAPH}, 2022.

\bibitem[Schuhmann et~al.(2021)Schuhmann, Vencu, Beaumont, Kaczmarczyk, Mullis, Katta, Coombes, Jitsev, and Komatsuzaki]{schuhmann2021laion}
Christoph Schuhmann, Richard Vencu, Romain Beaumont, Robert Kaczmarczyk, Clayton Mullis, Aarush Katta, Theo Coombes, Jenia Jitsev, and Aran Komatsuzaki.
\newblock Laion-400m: Open dataset of clip-filtered 400 million image-text pairs.
\newblock \emph{arXiv preprint arXiv:2111.02114}, 2021.

\bibitem[Shen et~al.(2015)Shen, Shen, Liu, and Tao~Shen]{shen2015supervised}
Fumin Shen, Chunhua Shen, Wei Liu, and Heng Tao~Shen.
\newblock Supervised discrete hashing.
\newblock In \emph{CVPR}, 2015.

\bibitem[Song et~al.(2021)Song, Meng, and Ermon]{songdenoising}
Jiaming Song, Chenlin Meng, and Stefano Ermon.
\newblock Denoising diffusion implicit models.
\newblock In \emph{ICLR}, 2021.

\bibitem[Sun et~al.(2023)Sun, Pan, Ge, Li, Duan, Wu, Zhang, Zhou, Qin, Wang, et~al.]{sun2024journeydb}
Keqiang Sun, Junting Pan, Yuying Ge, Hao Li, Haodong Duan, Xiaoshi Wu, Renrui Zhang, Aojun Zhou, Zipeng Qin, Yi~Wang, et~al.
\newblock Journeydb: A benchmark for generative image understanding.
\newblock In \emph{NeurIPS}, 2023.

\bibitem[Sun et~al.(2024)Sun, Jiang, Chen, Zhang, Peng, Luo, and Yuan]{sun2024autoregressive}
Peize Sun, Yi~Jiang, Shoufa Chen, Shilong Zhang, Bingyue Peng, Ping Luo, and Zehuan Yuan.
\newblock Autoregressive model beats diffusion: Llama for scalable image generation.
\newblock \emph{arXiv preprint arXiv:2406.06525}, 2024.

\bibitem[Tian et~al.(2024)Tian, Jiang, Yuan, Peng, and Wang]{tian2024visual}
Keyu Tian, Yi~Jiang, Zehuan Yuan, Bingyue Peng, and Liwei Wang.
\newblock Visual autoregressive modeling: Scalable image generation via next-scale prediction.
\newblock \emph{arXiv preprint arXiv:2404.02905}, 2024.

\bibitem[Touvron et~al.(2023{\natexlab{a}})Touvron, Lavril, Izacard, Martinet, Lachaux, Lacroix, Rozi{\`e}re, Goyal, Hambro, Azhar, et~al.]{touvron2023llama1}
Hugo Touvron, Thibaut Lavril, Gautier Izacard, Xavier Martinet, Marie-Anne Lachaux, Timoth{\'e}e Lacroix, Baptiste Rozi{\`e}re, Naman Goyal, Eric Hambro, Faisal Azhar, et~al.
\newblock Llama: Open and efficient foundation language models.
\newblock \emph{arXiv preprint arXiv:2302.13971}, 2023{\natexlab{a}}.

\bibitem[Touvron et~al.(2023{\natexlab{b}})Touvron, Martin, Stone, Albert, Almahairi, Babaei, Bashlykov, Batra, Bhargava, Bhosale, et~al.]{touvron2023llama2}
Hugo Touvron, Louis Martin, Kevin Stone, Peter Albert, Amjad Almahairi, Yasmine Babaei, Nikolay Bashlykov, Soumya Batra, Prajjwal Bhargava, Shruti Bhosale, et~al.
\newblock Llama 2: Open foundation and fine-tuned chat models.
\newblock \emph{arXiv preprint arXiv:2307.09288}, 2023{\natexlab{b}}.

\bibitem[Tschannen et~al.(2023)Tschannen, Eastwood, and Mentzer]{tschannen2023givt}
Michael Tschannen, Cian Eastwood, and Fabian Mentzer.
\newblock Givt: Generative infinite-vocabulary transformers.
\newblock \emph{arXiv preprint arXiv:2312.02116}, 2023.

\bibitem[Van Den~Oord et~al.(2017)Van Den~Oord, Vinyals, et~al.]{van2017neural}
Aaron Van Den~Oord, Oriol Vinyals, et~al.
\newblock Neural discrete representation learning.
\newblock In \emph{NeurIPS}, 2017.

\bibitem[van~der Maaten \& Hinton(2008)van~der Maaten and Hinton]{tsne}
Laurens van~der Maaten and Geoffrey Hinton.
\newblock Visualizing data using t-sne.
\newblock \emph{JMLR}, 2008.

\bibitem[Wang et~al.(2017)Wang, Zhang, Sebe, Shen, et~al.]{wang2017survey}
Jingdong Wang, Ting Zhang, Nicu Sebe, Heng~Tao Shen, et~al.
\newblock A survey on learning to hash.
\newblock \emph{IEEE TPAMI}, 2017.

\bibitem[Wang et~al.(2023)Wang, Wang, Liu, and Qiu]{wang2023binary}
Ze~Wang, Jiang Wang, Zicheng Liu, and Qiang Qiu.
\newblock Binary latent diffusion.
\newblock In \emph{CVPR}, 2023.

\bibitem[Wei et~al.(2021)Wei, Shen, Sun, Ye, and Yang]{wei20212}
Xiu-Shen Wei, Yang Shen, Xuhao Sun, Han-Jia Ye, and Jian Yang.
\newblock A$^2$-net: Learning attribute-aware hash codes for large-scale fine-grained image retrieval.
\newblock In \emph{NeurIPS}, 2021.

\bibitem[Wu et~al.(2019)Wu, Dai, Liu, Li, and Wang]{wu2019deep}
Dayan Wu, Qi~Dai, Jing Liu, Bo~Li, and Weiping Wang.
\newblock Deep incremental hashing network for efficient image retrieval.
\newblock In \emph{CVPR}, 2019.

\bibitem[Yu et~al.(2022{\natexlab{a}})Yu, Li, Koh, Zhang, Pang, Qin, Ku, Xu, Baldridge, and Wu]{yu2022vector}
Jiahui Yu, Xin Li, Jing~Yu Koh, Han Zhang, Ruoming Pang, James Qin, Alexander Ku, Yuanzhong Xu, Jason Baldridge, and Yonghui Wu.
\newblock Vector-quantized image modeling with improved vqgan.
\newblock In \emph{ICLR}, 2022{\natexlab{a}}.

\bibitem[Yu et~al.(2022{\natexlab{b}})Yu, Xu, Koh, Luong, Baid, Wang, Vasudevan, Ku, Yang, Ayan, Hutchinson, Han, Parekh, Li, Zhang, Baldridge, and Wu]{yu2022scaling}
Jiahui Yu, Yuanzhong Xu, Jing~Yu Koh, Thang Luong, Gunjan Baid, Zirui Wang, Vijay Vasudevan, Alexander Ku, Yinfei Yang, Burcu~Karagol Ayan, Ben Hutchinson, Wei Han, Zarana Parekh, Xin Li, Han Zhang, Jason Baldridge, and Yonghui Wu.
\newblock Scaling autoregressive models for content-rich text-to-image generation.
\newblock \emph{TMLR}, 2022{\natexlab{b}}.

\bibitem[Yu et~al.(2024)Yu, Lezama, Gundavarapu, Versari, Sohn, Minnen, Cheng, Gupta, Gu, Hauptmann, Gong, Yang, Essa, Ross, and Jiang]{yu2024language}
Lijun Yu, Jose Lezama, Nitesh~Bharadwaj Gundavarapu, Luca Versari, Kihyuk Sohn, David Minnen, Yong Cheng, Agrim Gupta, Xiuye Gu, Alexander~G Hauptmann, Boqing Gong, Ming-Hsuan Yang, Irfan Essa, David~A Ross, and Lu~Jiang.
\newblock Language model beats diffusion - tokenizer is key to visual generation.
\newblock In \emph{ICLR}, 2024.

\bibitem[Zhou et~al.(2022)Zhou, Wei, Wang, Shen, Xie, Yuille, and Kong]{zhou2022ibot}
Jinghao Zhou, Chen Wei, Huiyu Wang, Wei Shen, Cihang Xie, Alan Yuille, and Tao Kong.
\newblock Image {BERT} pre-training with online tokenizer.
\newblock In \emph{ICLR}, 2022.

\end{thebibliography}
\bibliographystyle{iclr2025_conference}

\newpage
\appendix
\section{Additional implementation details}
\label{appx:detail}
\textbf{Model configuration}\quad
The configuration settings for the model architecture, training and inference of \modelname across different model sizes are provided in~\Cref{tab:hyperparam}. Unless otherwise specified, our model uses this default setting in the main paper. \rebut{For the inference hyperparameters, we observe the following patterns: 
\begin{enumerate}
    \item \textbf{CFG scale}: A larger CFG scale produces smoother images with clearer class features, while a smaller scale enhances fine-grained details.
    \item \textbf{Gumbel temperature}: A higher Gumbel temperature increases generation diversity but reduces quality, whereas a lower temperature improves quality at the cost of diversity.
    \item \textbf{Sampling iterations}: 20–30 iterations generally perform well. Using 10 iterations speeds up generation but slightly reduces quality, while more than 30 iterations has minimal impact but slows down generation.
    \item \textbf{Diffusion timesteps}: 100 steps typically yield good results. Performance remains largely consistent across a broad range of 10–200 steps, with only marginal differences.
\end{enumerate}
}

\begin{table}[ht]
\centering
\captionof{table}{\textbf{The default configuration settings of three models: \modelname-L, \modelname-XL, \modelname-XXL.}}
\begin{minipage}[t]{0.32\textwidth}
    \centering
    \scalebox{0.8}{
    \begin{tabular}{c|c}
        \toprule
        \multicolumn{2}{c}{\modelname-L} \\ 
        \midrule
        Config & Value \\
        \midrule
        \multicolumn{2}{c}{Architecture} \\
        \midrule
        % \midrule
         Transformer layers & 24 \\
         Transformer heads & 16 \\
         Transformer dimensions & 1024 \\
         MLP layers & 3 \\
         MLP dimensions & 1024 \\
        \midrule
        \multicolumn{2}{c}{Training} \\
        \midrule
        Batch size & 1024 \\
        Epochs & 400 \\
        Weight decay & 2e-2 \\
        Learning rate & 1e-4 \\
        Total diffusion timesteps & 256 \\
        \midrule
        \multicolumn{2}{c}{Inference} \\
        \midrule
        CFG scale & 2.5 \\
        Sampling iterations & 20 \\
        Gumbel temperature & 0.17  \\
        Diffusion timesteps & 100 \\
        \bottomrule
    \end{tabular}}
\end{minipage}
\hfill
\begin{minipage}[t]{0.32\textwidth}
    \centering
    \scalebox{0.8}{
    \begin{tabular}{c|c}
        \toprule
        \multicolumn{2}{c}{\modelname-XL} \\ 
        \midrule
        Config & Value \\
        \midrule
        \multicolumn{2}{c}{Architecture} \\
        \midrule
        % \midrule
         Transformer layers & 36 \\
         Transformer heads & 20 \\
         Transformer dimensions & 1280 \\
         MLP layers & 6 \\
         MLP dimensions & 1280 \\
        \midrule
        \multicolumn{2}{c}{Training} \\
        \midrule
        Batch size & 512 \\
        Epochs & 400 \\
        Weight decay & 2e-2 \\
        Learning rate & 1e-4 \\
        Total diffusion timesteps & 256 \\
        \midrule
        \multicolumn{2}{c}{Inference} \\
        \midrule
        CFG scale & 2.5 \\
        Sampling iterations & 25 \\
        Gumbel temperature & 0.25  \\
        Diffusion timesteps & 100 \\
        \bottomrule
    \end{tabular}}
\end{minipage}
\hfill
\begin{minipage}[t]{0.32\textwidth}
    \centering
    \scalebox{0.8}{
    \begin{tabular}{c|c}
        \toprule
        \multicolumn{2}{c}{\modelname-XXL} \\ 
        \midrule
        Config & Value \\
        \midrule
        \multicolumn{2}{c}{Architecture} \\
        \midrule
        % \midrule
         Transformer layers & 48 \\
         Transformer heads & 24 \\
         Transformer dimensions & 1536 \\
         MLP layers & 8 \\
         MLP dimensions & 1536 \\
        \midrule
        \multicolumn{2}{c}{Training} \\
        \midrule
        Batch size & 512 \\
        Epochs & 400 \\
        Weight decay & 2e-2 \\
        Learning rate & 1e-4 \\
        Total diffusion timesteps & 256 \\
        \midrule
        \multicolumn{2}{c}{Inference} \\
        \midrule
        CFG scale & 2.5 \\
        Sampling iterations & 25 \\
        Gumbel temperature & 0.30  \\
        Diffusion timesteps & 100 \\
        \bottomrule
    \end{tabular}}
\end{minipage}
\label{tab:hyperparam}
\end{table}

\textbf{Binary transcoder}\quad
After producing Bernoulli distribution probabilities through Bernoulli denoising, there are two ways to obtain binary codes: deterministic and non-deterministic. For deterministic method, values are set to $1$ if the probability exceeds $0.5$, and $0$ otherwise. In contrast, for non-deterministic methods, we sample directly from the Bernoulli distribution to obtain $0$ and $1$ values. We empirically compare these two methods, as shown in~\Cref{tab:deterministic} and find that the non-deterministic approach slightly outperforms its deterministic counterpart. 
As a result, we adopt the non-deterministic approach for all models presented in the main paper.

\begin{table}[ht]
   \centering
    \caption{\textbf{Comparison of deterministic and non-deterministic sampling.}} 
    \scalebox{1.0}{
    \begin{tabular}{lccccc}
        \toprule
        Determ. & FID$\downarrow$ & IS$\uparrow$ & sFID$\downarrow$ & Pre.$\uparrow$ & Rec.$\uparrow$ \\
        \midrule
        \cmark & 3.19 & 239.79 & 6.25 & 0.84 & \textbf{0.52} \\
        \xmark\ (Ours) & \textbf{3.17} & \textbf{262.14} & \textbf{5.59} & \textbf{0.86} & 0.50 \\
        \bottomrule
    \end{tabular}}
    \label{tab:deterministic} 
\end{table}

\textbf{Sampling strategy}\quad
In our sampling strategy, the implementation of \Cref{eq:bin_ent} in the main paper may encounter a "nan" issue caused by the logarithmic operation. Since we only need to compare the relative magnitudes of different entries, we can instead use a value with the same trend to mimic the exact confidence. We use $2\times \lvert p_k - 0.5 \rvert$ as the final confidence value in our implementation.

\textbf{Adaptive LayerNorm}\quad
We empirically find that adaptive LayerNorm (adaLN) has a marginal effect on performance. Following the approach in~\citet{tian2024visual}, we implement a shared adaLN that uses a single MLP to obtain the shift, scale, and gate values for all transformer layers, which adds only a minimal number of parameters.

\textbf{Linear probe}\quad
Following the linear-probe evaluation protocol outlined in~\citep{he2022masked}, we use the LARS optimizer with a momentum of $0.9$. We train the linear head for 100 epochs, using a batch size of 256 along with 8 gradient accumulation steps. We use a warm-up period of 10 epochs and set the learning rate to 0.1. An extra BatchNorm layer is added before the linear classifier, without affine transformation. We refrain from using mixup, cutmix, drop path, or color jittering, and the weight decay is set to zero. We use the same linear-probe setting for all compared models in the main paper.

\section{\rebut{Text-to-image generation}}
\rebut{We validate the effectiveness of BiGR for text-to-image generation (T2I).}

\rebut{
\textbf{Model adaptation}\quad
We follow the practice of LlamaGen to adapt our class-conditional image generation model into a text-to-image generation model. Most parts of our model remain unchanged, except for replacing the class-condition token embedding with the text token embedding extracted by a text encoder. 
We use T5-XXL~\citep{raffel2020exploring}\footnote{\url{https://huggingface.co/google/t5-v1_1-xxl}} as our text encoder. The maximum text token length is set to 120, with left-padding applied during both training and inference. 
The extracted text token embeddings are projected through an additional MLP layer and appended to the patch tokens. Conditioning for image generation relies solely on these appended text token embeddings, without using adaptive LayerNorm or cross-attention. Note that this minimal model adaptation is intended to validate the model's capability for text-to-image generation. More refined designs could be implemented to enhance generation performance, which we leave for future research.
}

\rebut{
\textbf{Training dataset}\quad
We train our T2I model on a 20M subset of LAION-400M~\citep{schuhmann2021laion} and JourneyDB~\citep{sun2024journeydb}, a large-scale dataset containing 4M high-quality images annotated with corresponding text captions. The training images are center-cropped and resized to resolutions of 256$\times$256 and 512$\times$512.
}

\rebut{
\textbf{Implementation details}\quad
We conduct experiments using BiGR-XL-d24. For simplicity, we train our T2I model in a single stage, unlike the two-stage approach used in LlamaGen. The T2I model is initialized with the pretrained weights of our class-conditional generation model. 
We train the model on 32 A800 GPUs in three stages: (1) We initially train the model for 250K steps with a batch size of 1024 using JourneyDB at a resolution of 256$\times$256; (2) We then finetune it for 450K steps with a batch size of 256 on both datasets at a resolution of 512$\times$512; (3) Finally, we further finetune the model for 500K steps on JourneyDB at a resolution of 512$\times$512.
}

\begin{figure}[p]
    \centering
    \includegraphics[width=1.0\linewidth]{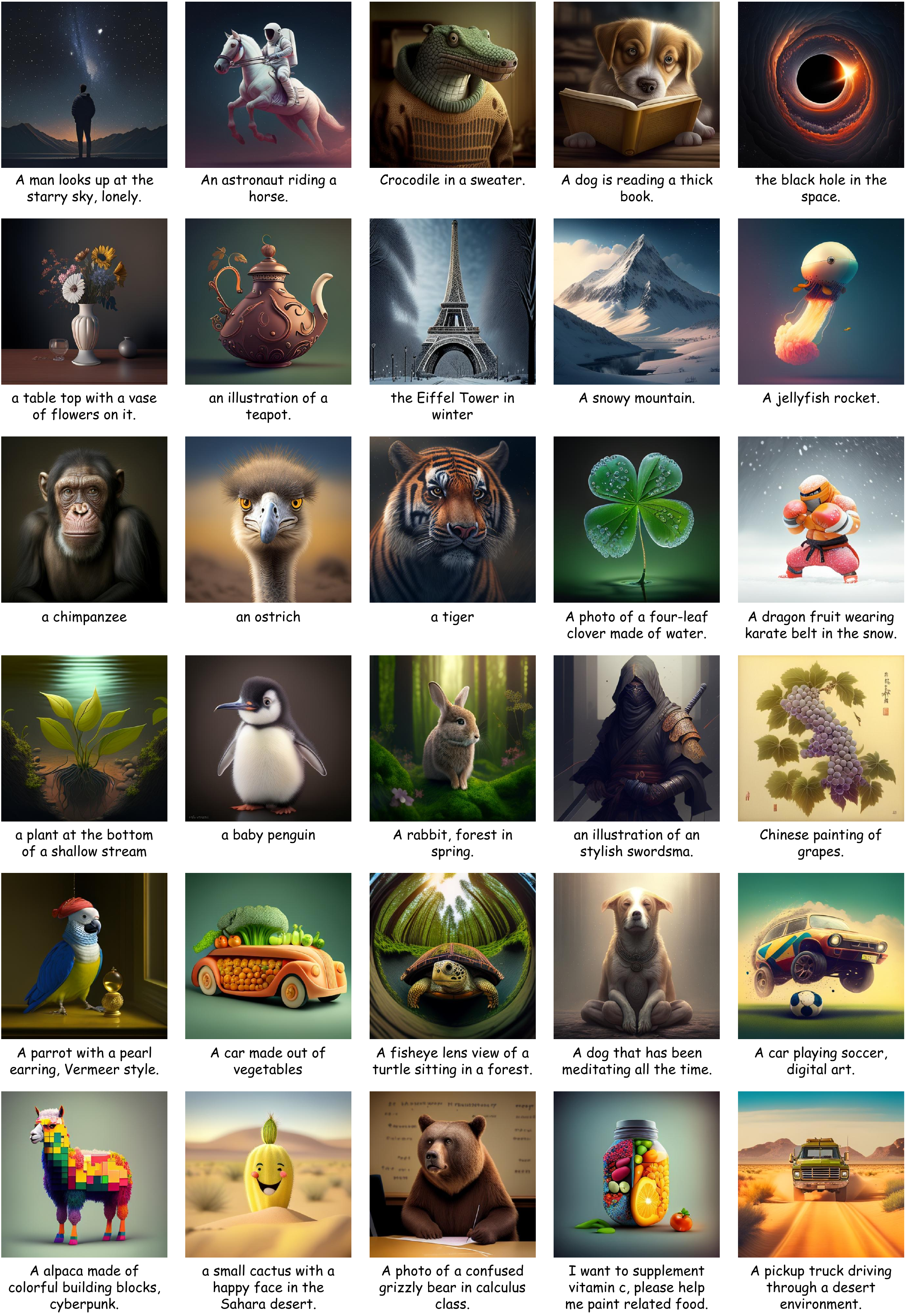}
    \caption{\rebut{\textbf{Text-conditional 512$\times$512 image generation samples with short prompts.} The results are generated by BiGR-XL-d24.}}
    \label{fig:t2i-short}
\end{figure}

\begin{figure}[p]
    \centering
    \includegraphics[width=1.0\linewidth]{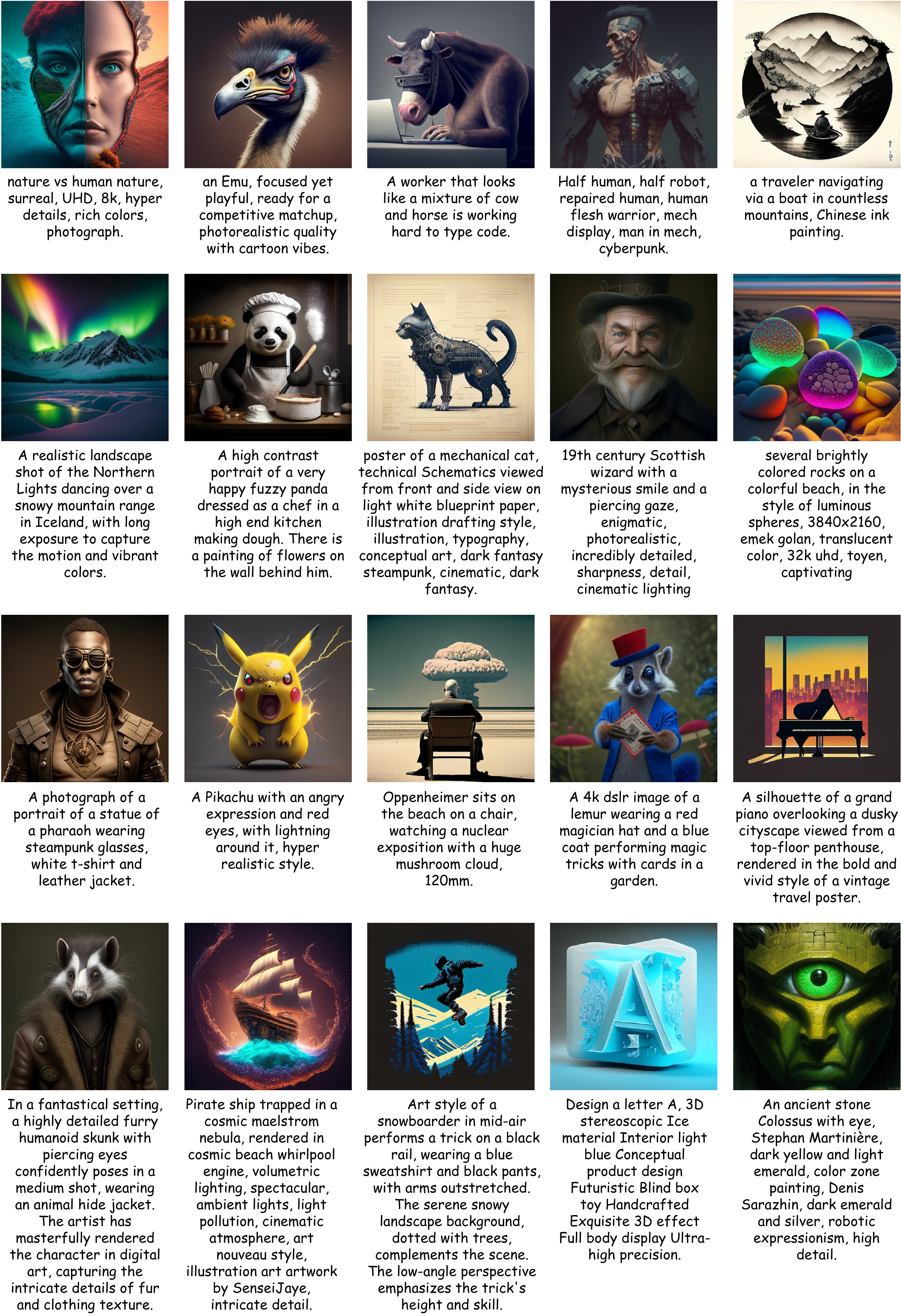}
    \caption{\rebut{\textbf{Text-conditional 512$\times$512 image generation samples with long prompts.} The results are generated by BiGR-XL-d24.}}
    \label{fig:t2i-long}
\end{figure}

\rebut{
\textbf{Generation samples}\quad
We present our text-to-image generation samples with short prompts in~\Cref{fig:t2i-short} and with long prompts in~\Cref{fig:t2i-long}, showing that our model performs well with both short and long prompts. The prompts we use are selected from LlamaGen~\citep{sun2024autoregressive} and various T2I works, such as PixArt~\citep{chenpixart}, Imagen~\citep{saharia2022imagen}, and DALL-E~3~\citep{betker2023dalle3}. These prompts are unseen in the training set, demonstrating that our T2I model generalizes well. 
Current results reveal the strong potential of BiGR in text-to-image generation.
Note that our training data, model size, and training duration are relatively limited, and we only train a 512$\times$512 T2I model, which has a relatively low image resolution.
We believe that scaling up training data and model size, along with exploring advanced techniques such as incorporating higher resolutions, fine-tuning autoencoders, and using images with higher quality, can further enhance BiGR's T2I performance. We will explore this in future research.
}

\section{\rebut{Visualization of the sampling order}}
\begin{figure}[t]
    \centering
    \includegraphics[width=0.95\linewidth]{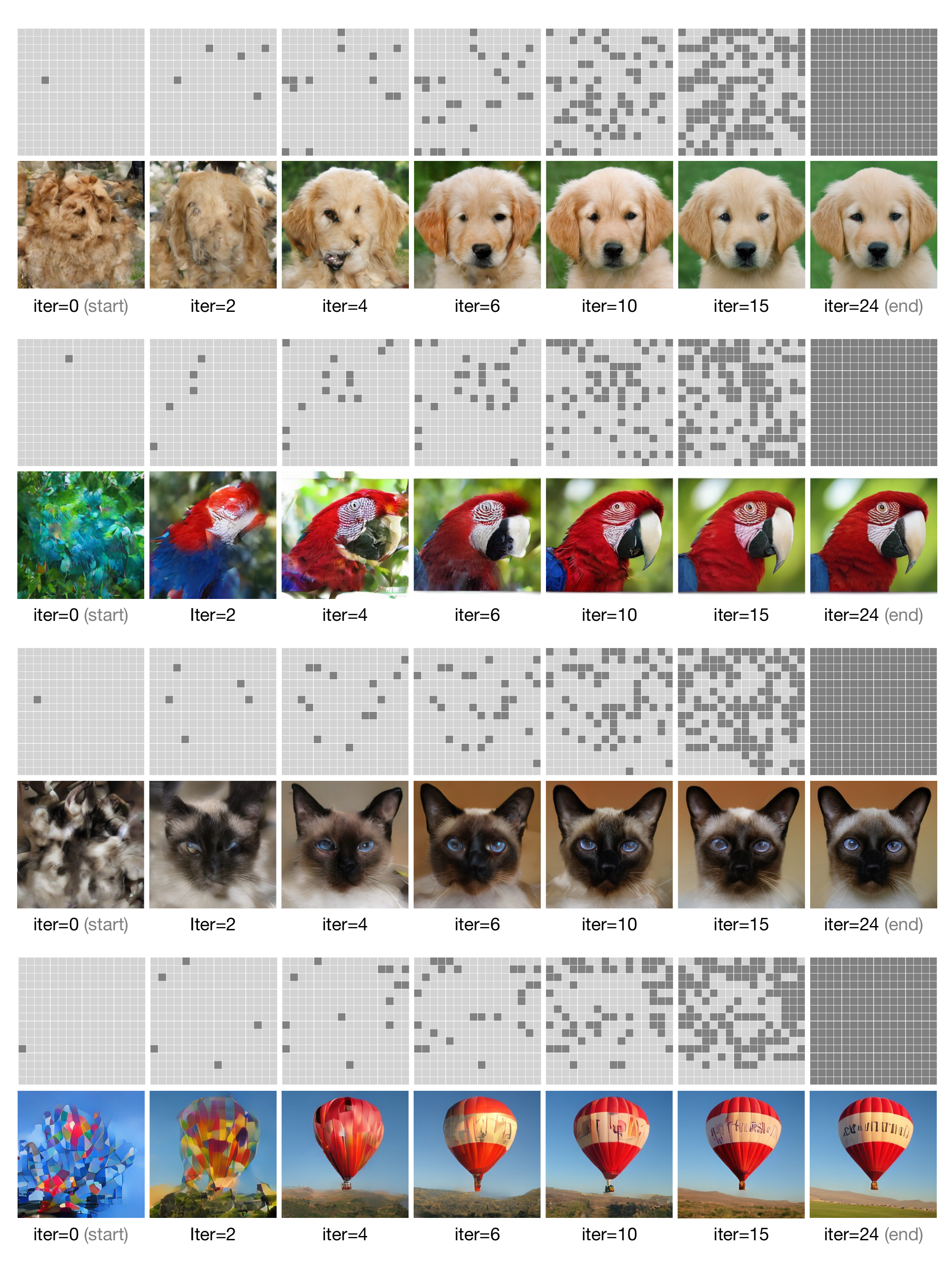}
    \caption{\rebut{\textbf{Visualization of the entropy-ordered sampling process.} For each generation sample, we present the illustration of the unmasked tokens at the top and the corresponding generated images at the bottom for each sampling iteration. At each iteration, the masked positions are filled with the output tokens from the current iteration. Here, we use BiGR-XL-d24, setting the total number of sampling iterations to 25.}}
    \label{fig:sample-order}
\end{figure}

\rebut{
We propose an entropy-ordered sampling process in this paper. To explore this sampling order further, we visualize the generated results at different iterations within the process. The visualization of the process is presented in~\Cref{fig:sample-order}.
We observe that early iterations capture class-level characteristics, while subsequent iterations generate finer object-related details. In the final stages, visual quality steadily improves.
}

\section{\rebut{Failure cases}}
\begin{figure}[t]
    \centering
    \includegraphics[width=0.95\linewidth]{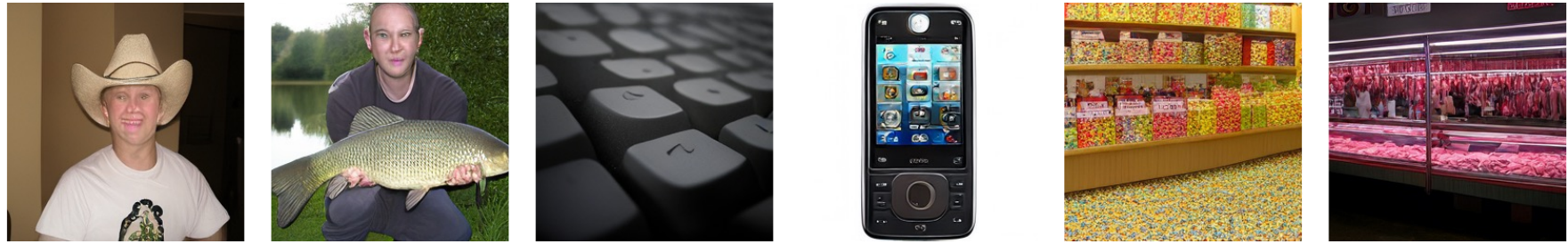}
    \caption{\rebut{\textbf{Failure cases} in generating high-quality human faces and fingers, clear numbers and signs, and complex scenes.}}
    \label{fig:fail}
\end{figure}

\rebut{We show some failure cases of our generation results in~\Cref{fig:fail}. 
It is challenging to generate high-quality, authentic human faces and fingers, clear and accurate numbers and signs, and highly complex scenes with many objects.
These issues are common in conditional generative models, arising from the significant visual quality differences across categories in the ImageNet-1K training dataset and the autoencoder's limited ability to reconstruct complex details.}

\section{Additional system-level comparison}
\label{appx:compare-more}
We provide a more comprehensive comparison of different leading models. We compare generative performance in~\Cref{appx:compare-gen} and discriminative performance in~\Cref{appx:compare-linprobe}.

\section{Additional generated samples}
\label{appx:more}
\begin{figure}[t]
    \centering
    \includegraphics[width=0.95\linewidth]{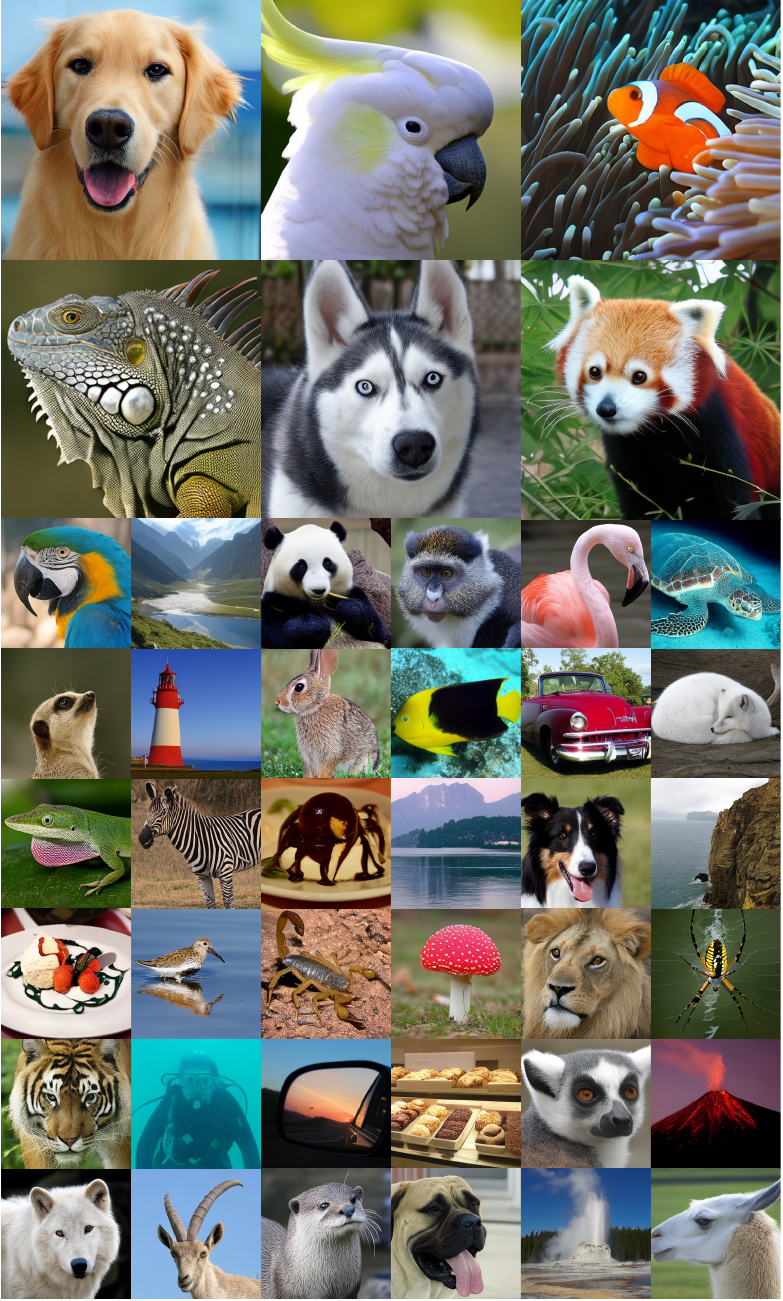}
    \caption{\textbf{Additional generated 256$\times$256 and 512$\times$512 samples.}}
    \label{fig:more}
\end{figure}

\begin{figure}[htbp]
\centering
\begin{minipage}{0.49\linewidth}
\centering
\includegraphics[width=1.0\linewidth]{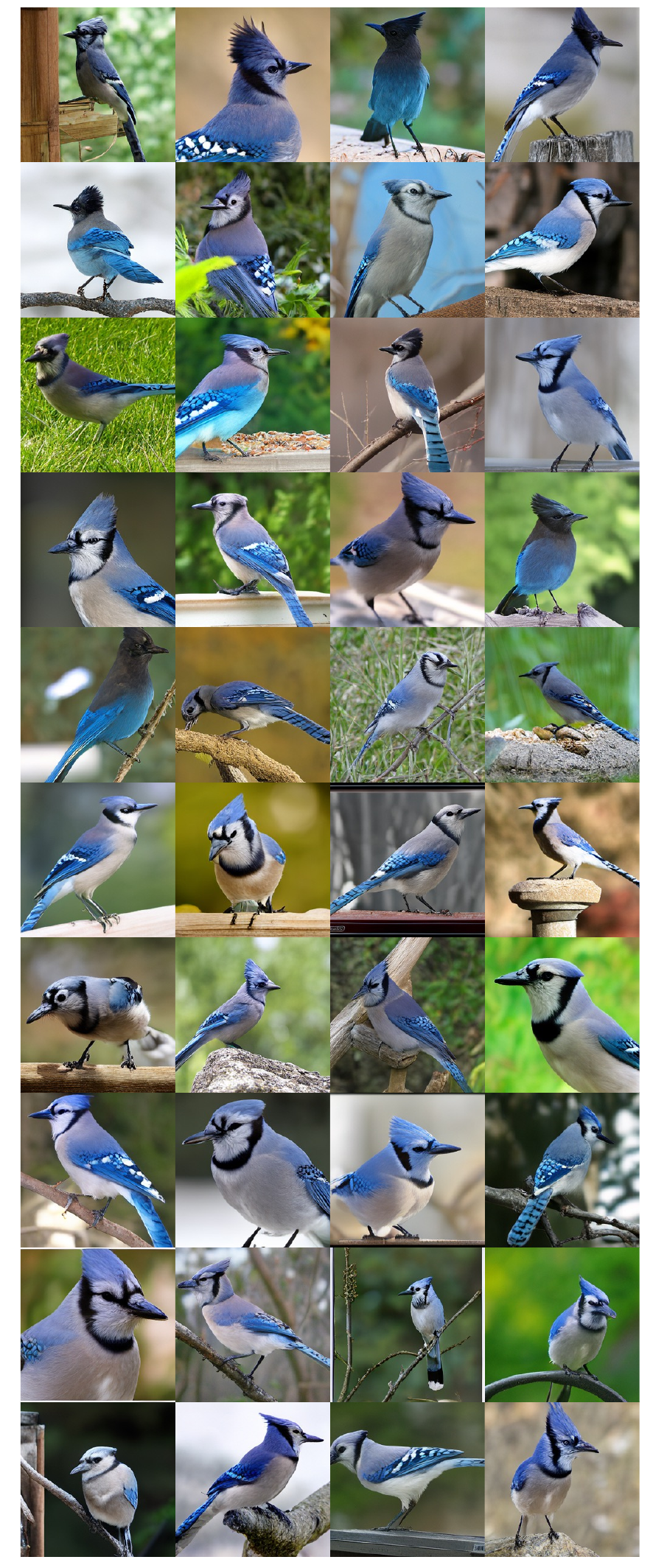}
\caption{\textbf{Uncurated} 256$\times$256 samples.\\
Model: \modelname-XXL-d32\\
Class label = “Jay” (17)\\}
\label{cls1}
\end{minipage}
%\qquad
\begin{minipage}{0.49\linewidth}
\centering
\includegraphics[width=1.0\linewidth]{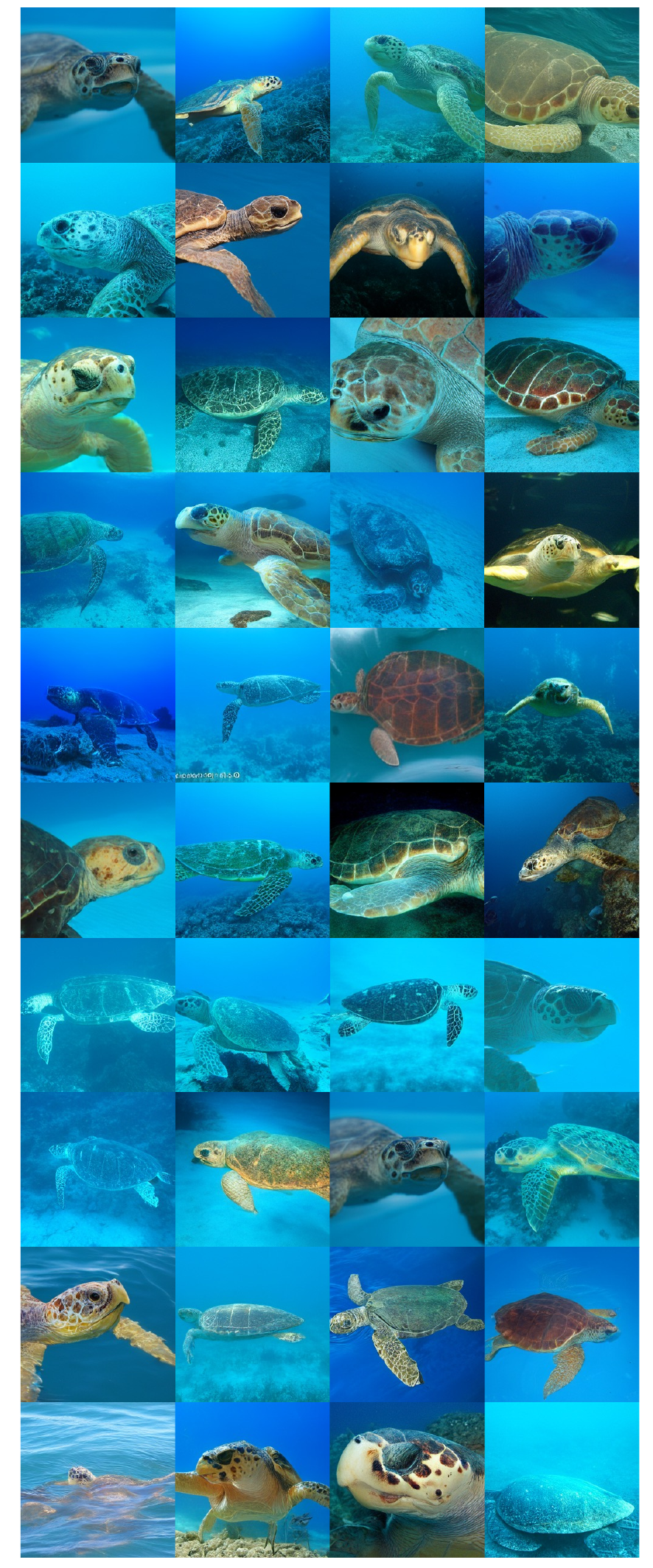}
\caption{\textbf{Uncurated} 256$\times$256  samples.\\
Model: \modelname-XXL-d32\\
Class label = “Loggerhead, loggerhead turtle, Caretta caretta” (33)}
\label{cls2}
\end{minipage}
\end{figure}

\begin{figure}[htbp]
\centering
\begin{minipage}{0.49\linewidth}
\centering
\includegraphics[width=1.0\linewidth]{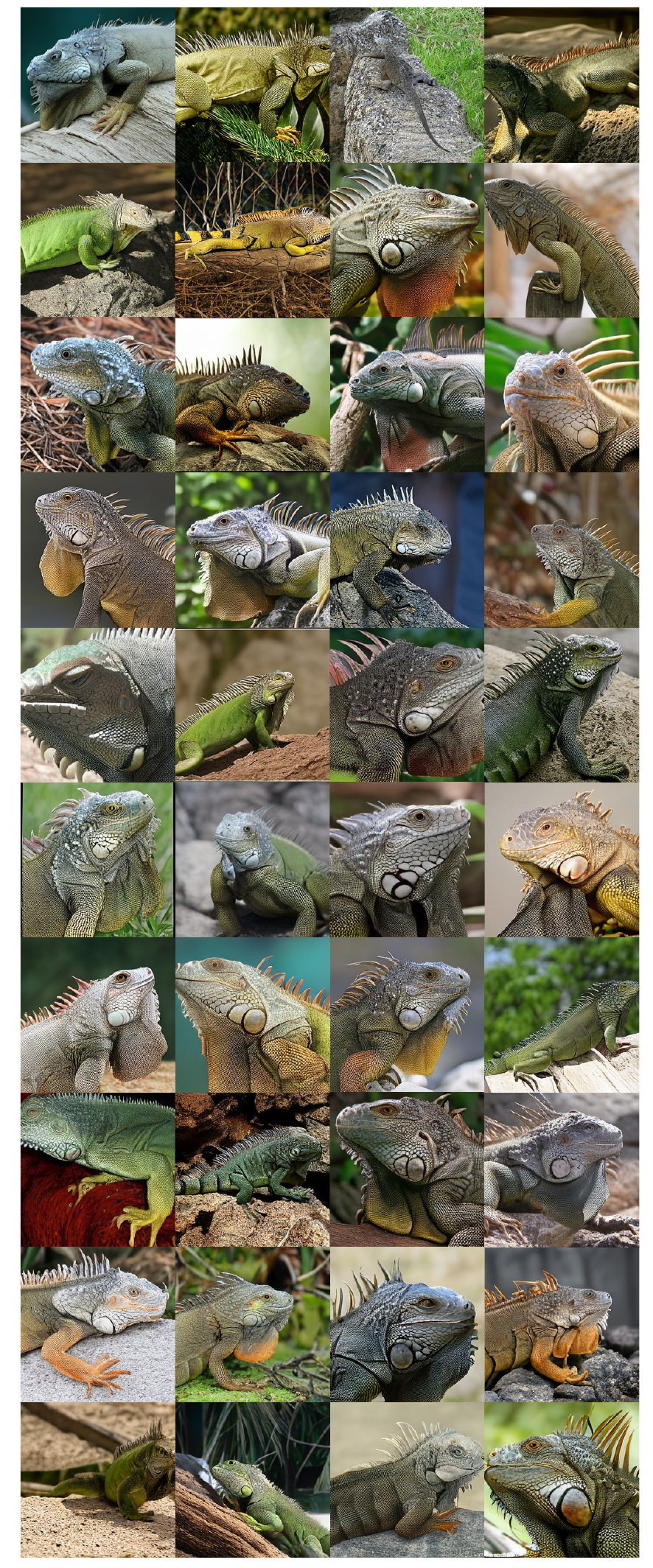}
\caption{\textbf{Uncurated} 256$\times$256 samples.\\
Model: \modelname-XXL-d32 \\
Class label = “Common iguana, Iguana, Igua-\\na iguana” (39)}
\label{cls3}
\end{minipage}
%\qquad
\begin{minipage}{0.49\linewidth}
\centering
\includegraphics[width=1.0\linewidth]{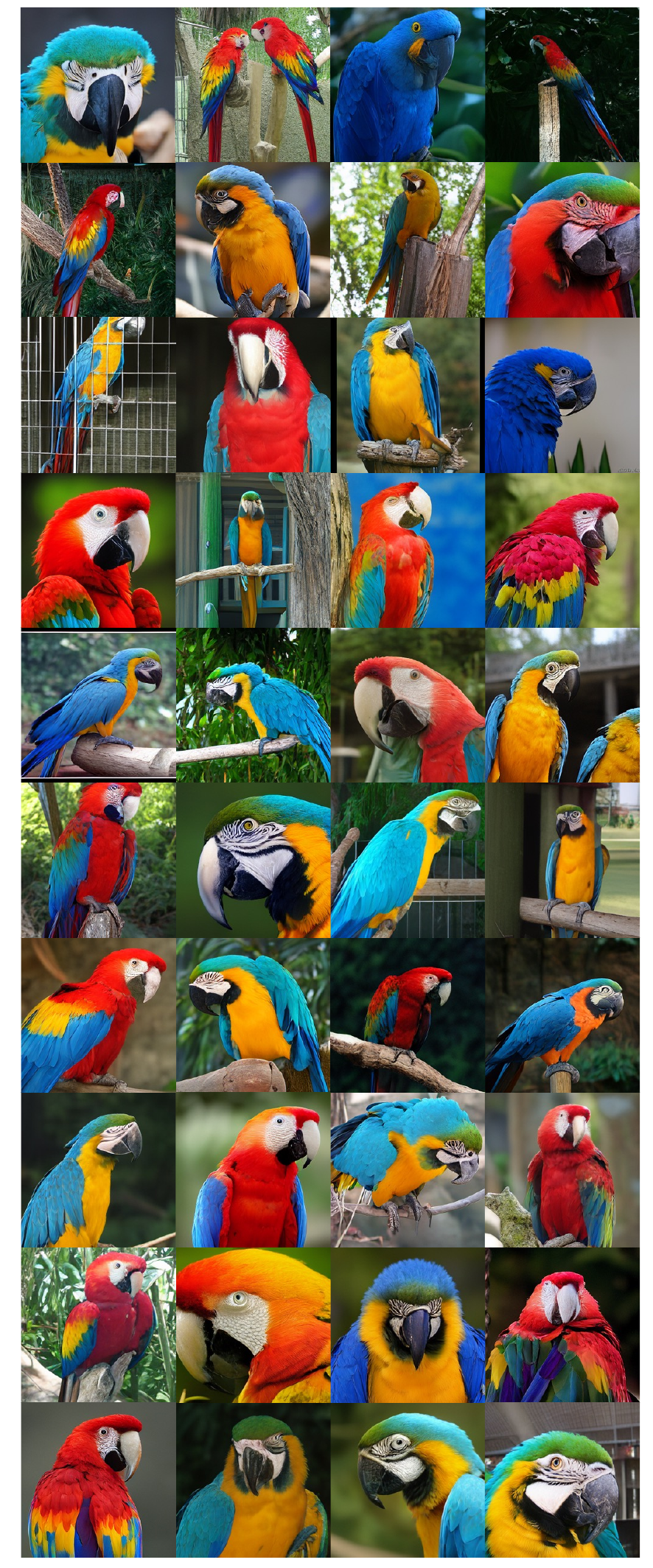}
\caption{\textbf{Uncurated} 256$\times$256 samples.\\
Model: \modelname-XXL-d32 \\
Class label = “Macaw” (88)\\}
\label{cls4}
\end{minipage}
\end{figure}

\begin{figure}[htbp]
\centering
\begin{minipage}{0.49\linewidth}
\centering
\includegraphics[width=1.0\linewidth]{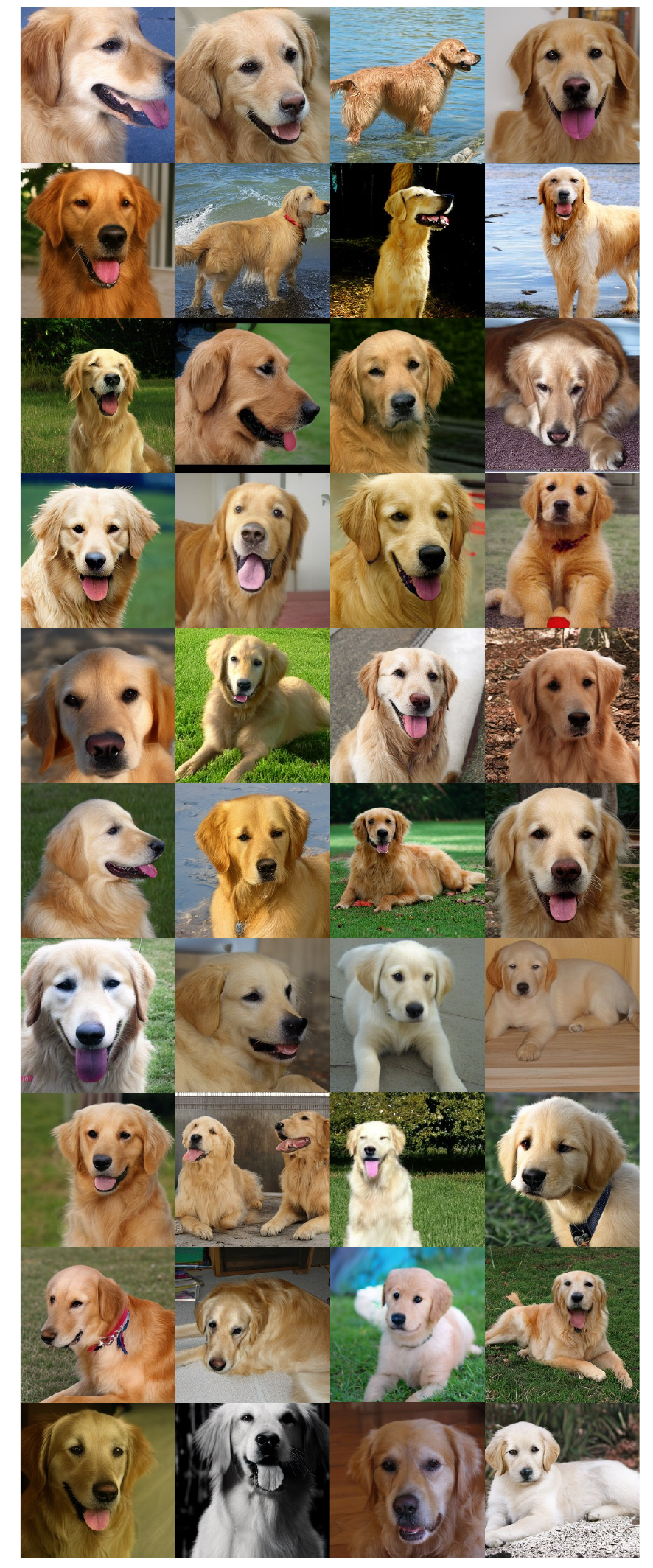}
\caption{\textbf{Uncurated} 256$\times$256  samples.\\
Model: \modelname-XXL-d32 \\
Class label = “Golden retriever” (207)}
\label{cls5}
\end{minipage}
%\qquad
\begin{minipage}{0.49\linewidth}
\centering
\includegraphics[width=1.0\linewidth]{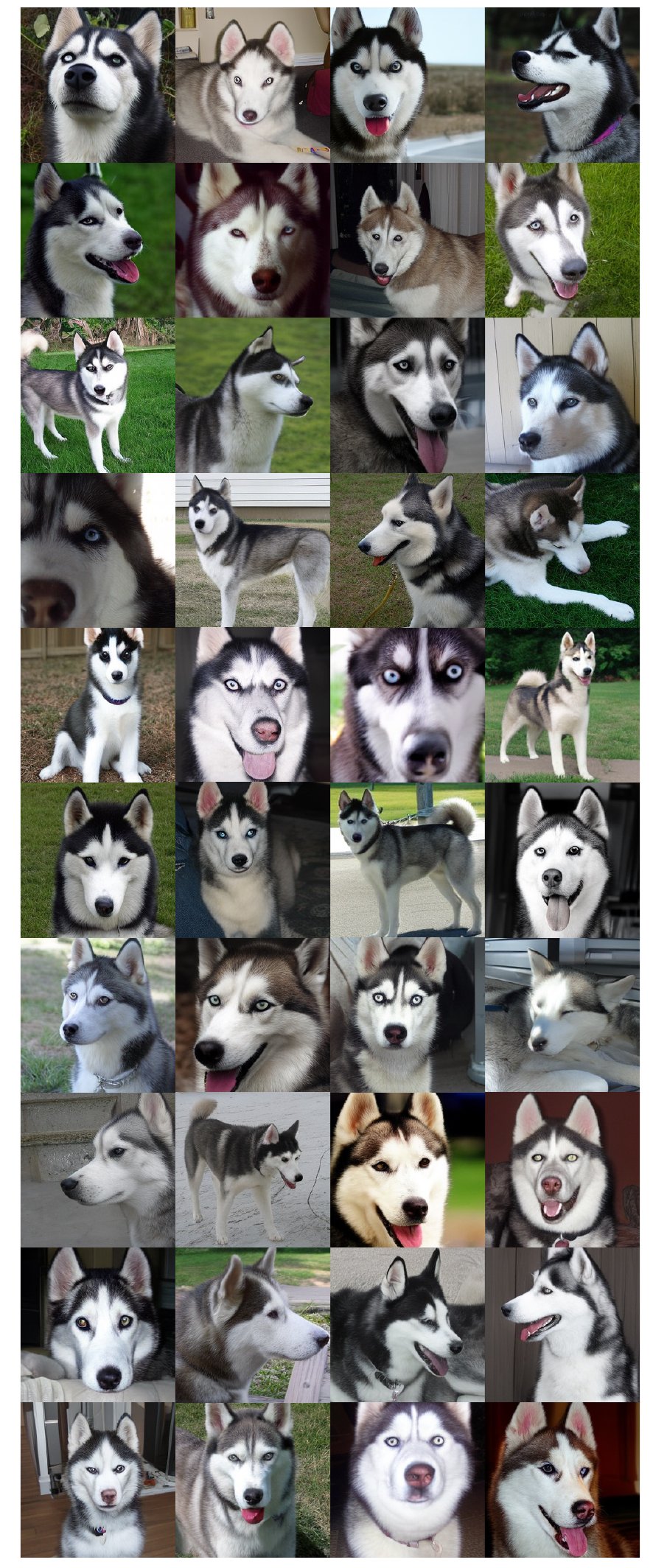}
\caption{\textbf{Uncurated} 256$\times$256  samples.\\
Model: \modelname-XXL-d32 \\
Class label = “Siberian husky” (250)}
\label{cls6}
\end{minipage}
\end{figure}

\begin{figure}[htbp]
\centering
\begin{minipage}{0.49\linewidth}
\centering
\includegraphics[width=1.0\linewidth]{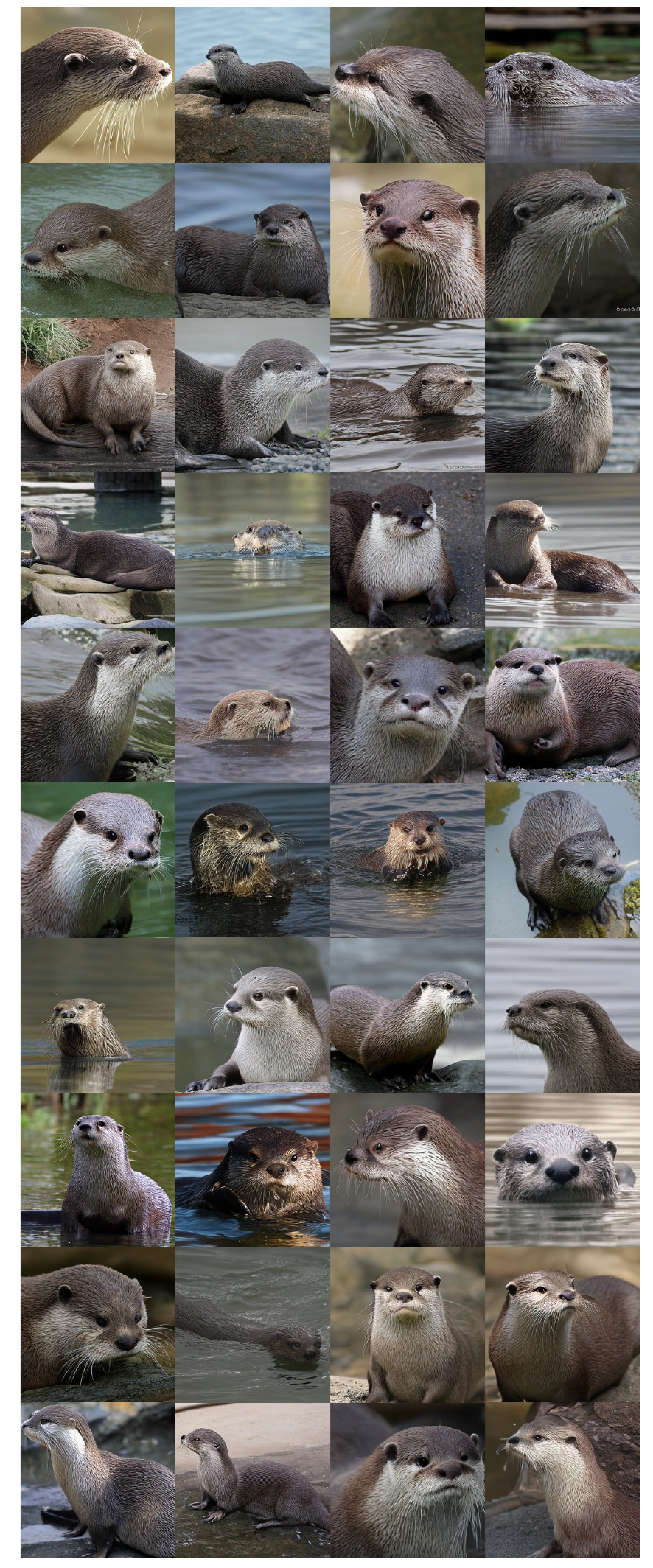}
\caption{\textbf{Uncurated} 256$\times$256  samples.\\
Model: \modelname-XXL-d32 \\
Class label = “Otter” (360)\\}
\label{cls7}
\end{minipage}
%\qquad
\begin{minipage}{0.49\linewidth}
\centering
\includegraphics[width=1.0\linewidth]{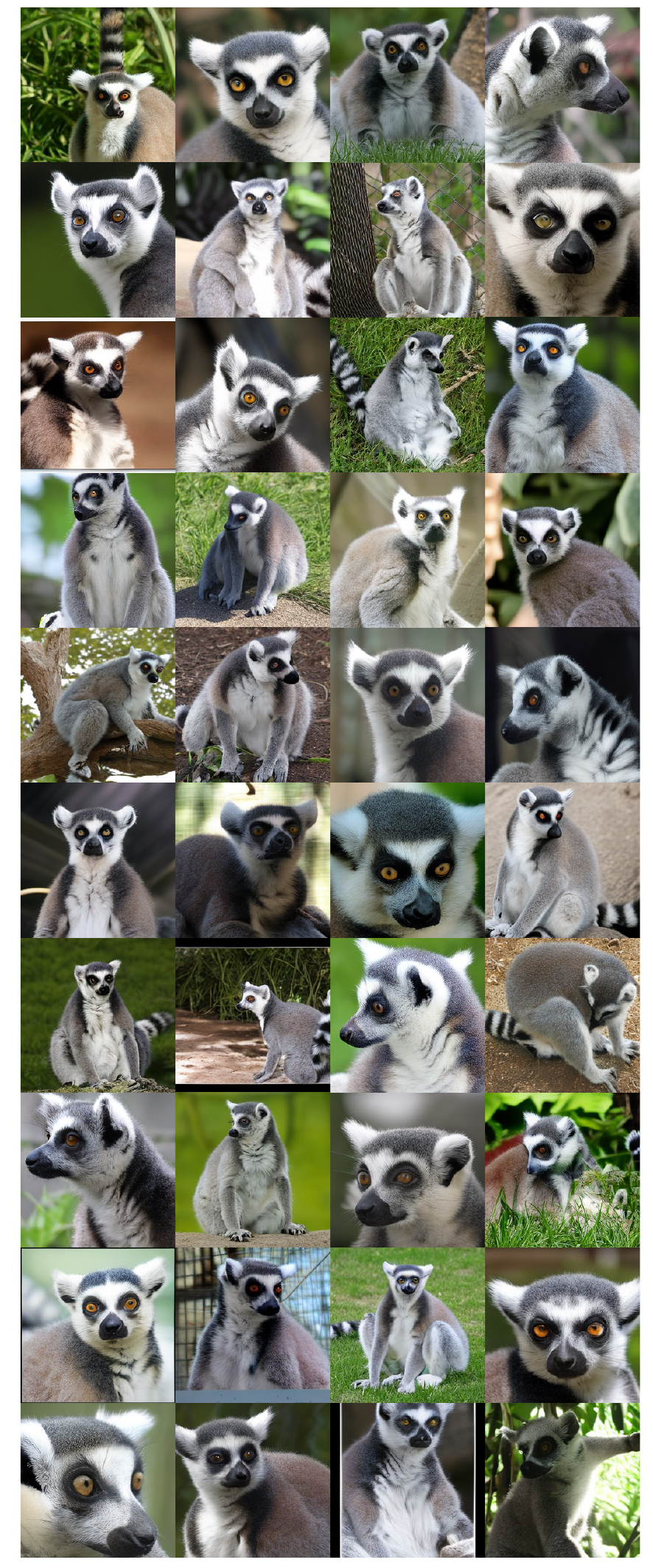}
\caption{\textbf{Uncurated} 256$\times$256  samples.\\
Model: \modelname-XXL-d32 \\
Class label = “Madagascar cat, ring-tailed lemur, Lemur catta” (383)}
\label{cls8}
\end{minipage}
\end{figure}

\begin{figure}[htbp]
\centering
\begin{minipage}{0.49\linewidth}
\centering
\includegraphics[width=1.0\linewidth]{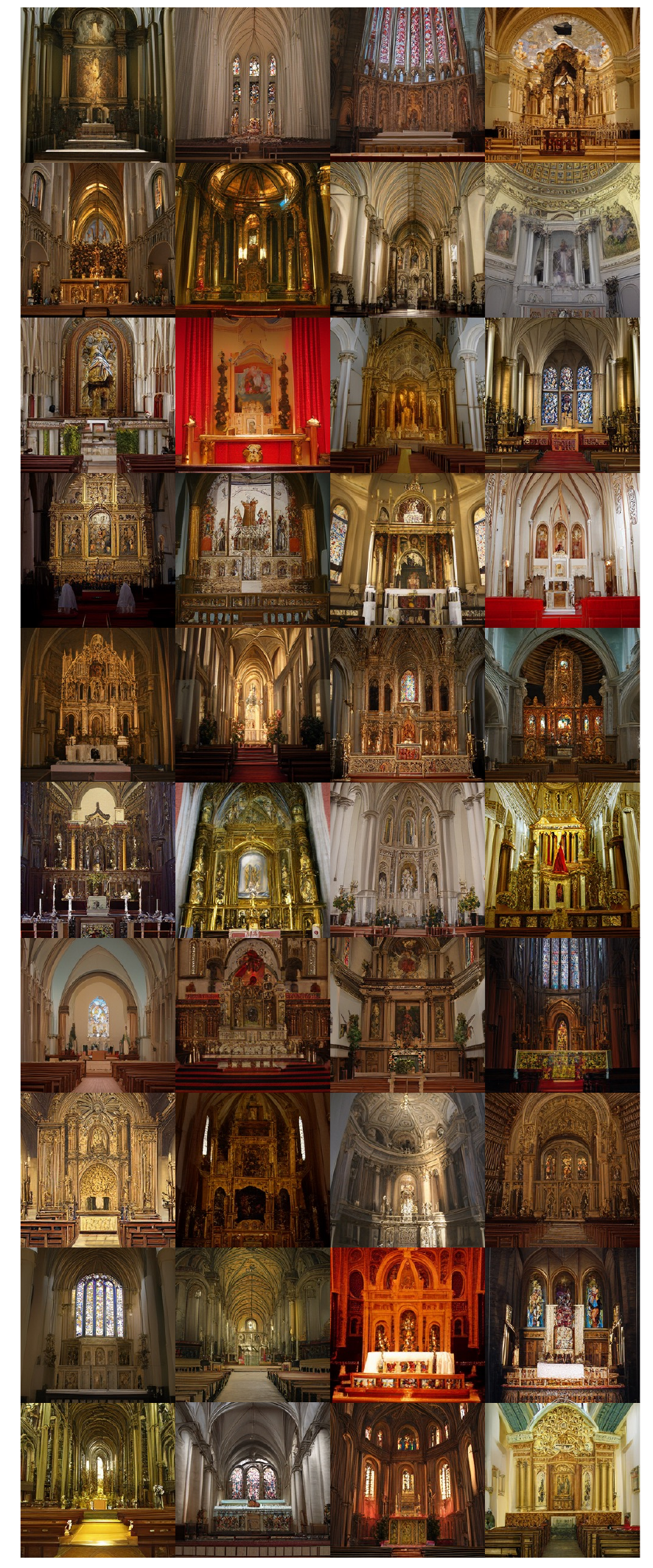}
\caption{\textbf{Uncurated} 256$\times$256  samples.\\
Model: \modelname-XXL-d32 \\
Class label = “Altar” (406)}
\label{cls9}
\end{minipage}
%\qquad
\begin{minipage}{0.49\linewidth}
\centering
\includegraphics[width=1.0\linewidth]{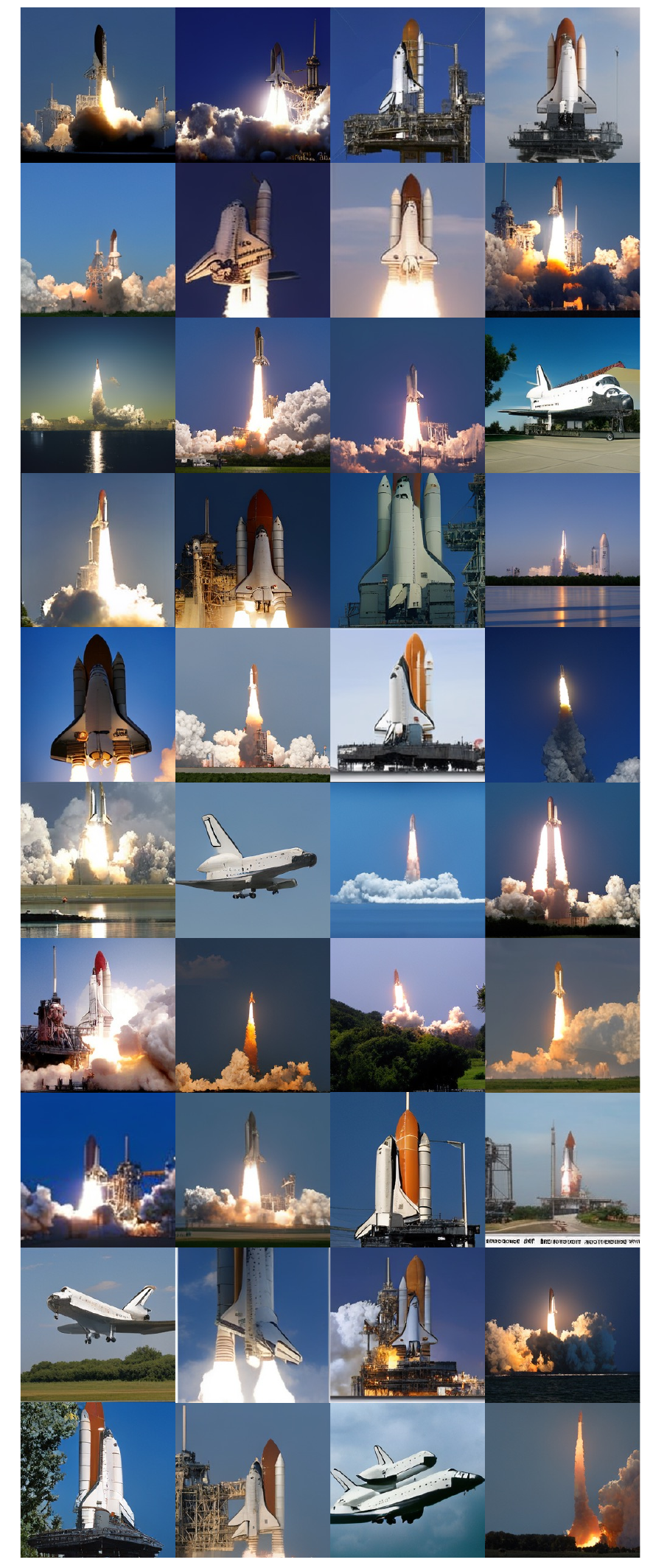}
\caption{\textbf{Uncurated} 256$\times$256  samples.\\
Model: \modelname-XXL-d32 \\
Class label = “Space Shuttle” (812)}
\label{cls10}
\end{minipage}
\end{figure}

\begin{figure}[htbp]
\centering
\begin{minipage}{0.49\linewidth}
\centering
\includegraphics[width=1.0\linewidth]{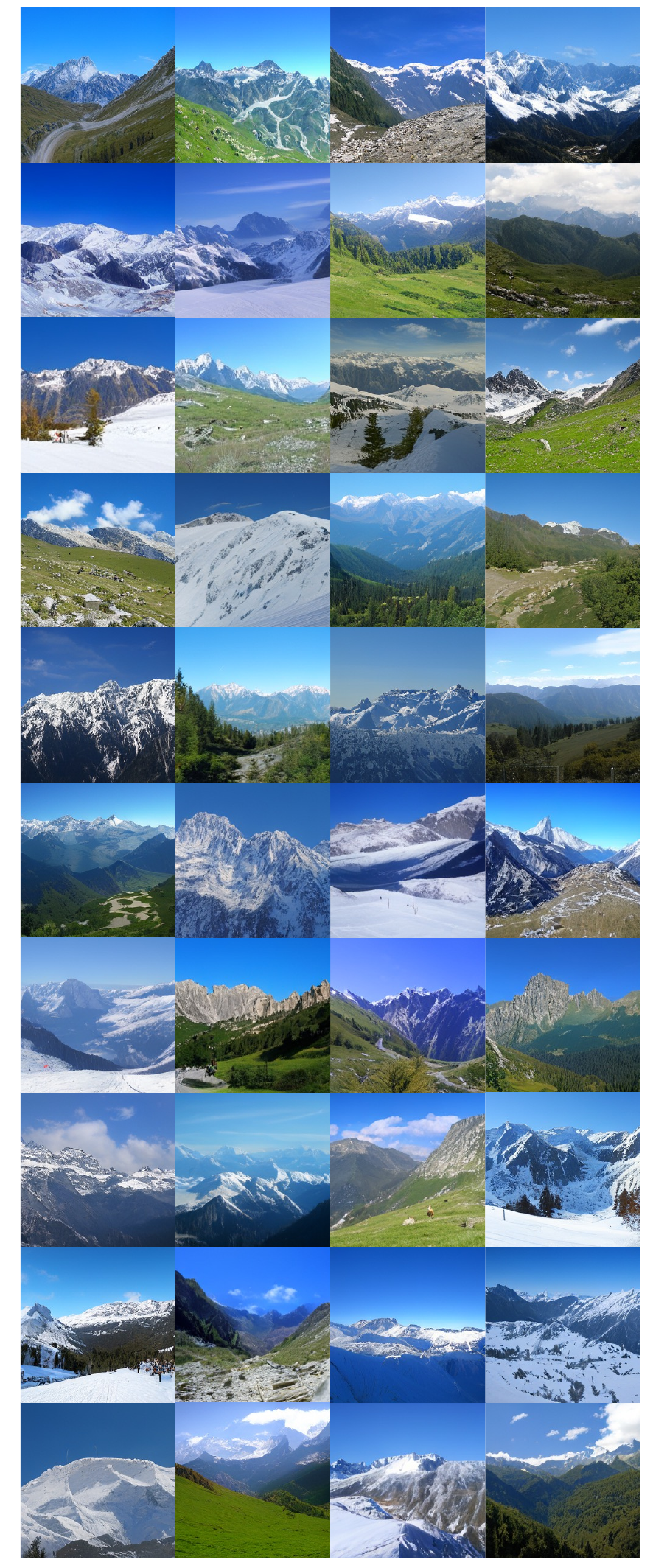}
\caption{\textbf{Uncurated} 256$\times$256  samples.\\
Model: \modelname-XXL-d32 \\
Class label = “Alp” (970)}
\label{cls11}
\end{minipage}
%\qquad
\begin{minipage}{0.49\linewidth}
\centering
\includegraphics[width=1.0\linewidth]{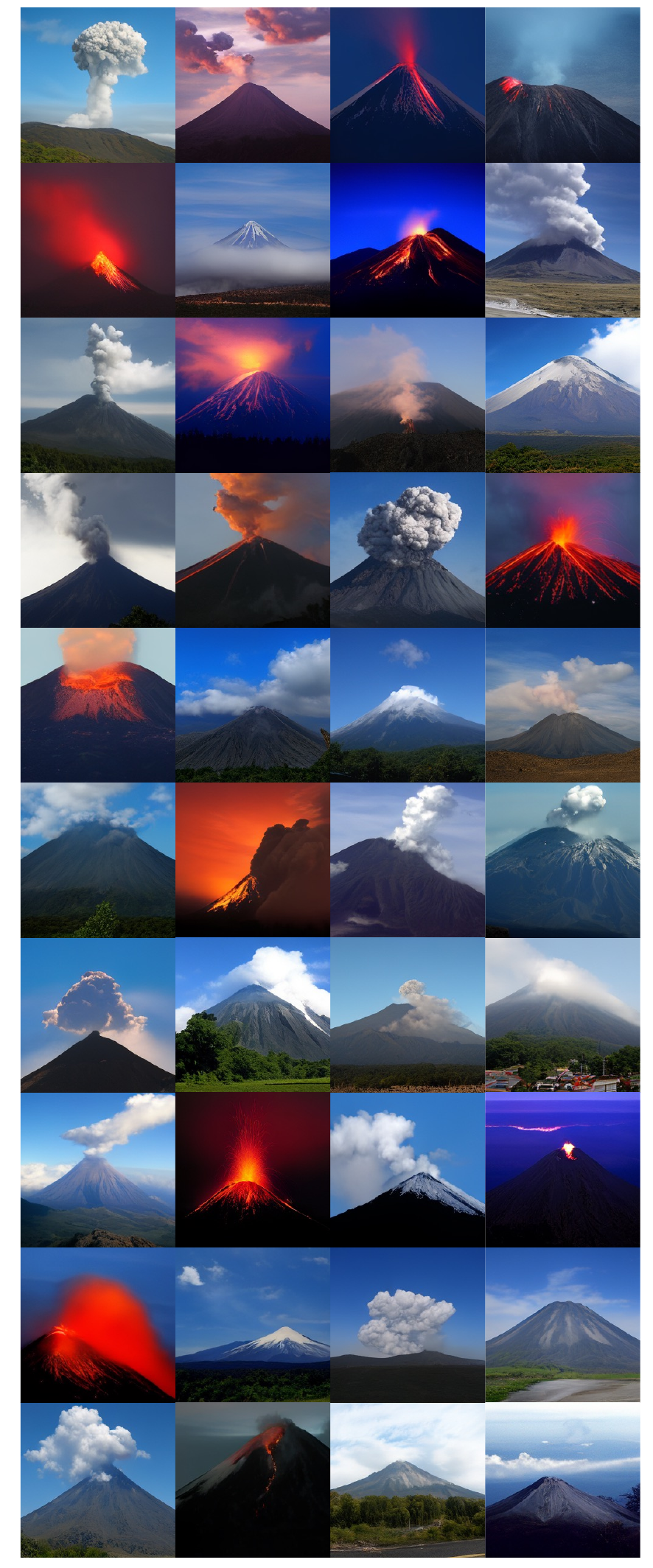}
\caption{\textbf{Uncurated} 256$\times$256  samples.\\
Model: \modelname-XXL-d32 \\
Class label = “Volcano” (980)}
\label{cls12}
\end{minipage}
\end{figure}

We provide additional 512$\times$512 samples and 256$\times$256 samples generated by our model in~\Cref{fig:more}. We also include uncurated generated samples from various classes in~\Cref{cls1,cls2,cls3,cls4,cls5,cls6,cls7,cls8,cls9,cls10,cls11,cls12}.

\section{Ethics statement}
We recognize the ethical risks of image generation, such as potential misuse for harmful content. Our research aims to promote positive uses like creativity and education, with a commitment to responsible application. Safeguards and continuous ethical oversight are strongly encouraged.

\begin{table}[t]
\centering
\caption{\textbf{Model comparison of generative performance on ImageNet-1K.}
Metrics include Frechet inception distance (FID), inception score (IS), precision (Pre.) and recall (Rec.). All models are tested on 256$\times$256 ImageNet-1K benchmark.
The suffix "-re" denotes the use of rejection sampling.}
\label{appx:compare-gen}
\scalebox{0.95}{
\begin{tabular}{ll|c|cccc}
\toprule
\textbf{Type} & \textbf{Model} & \textbf{\#Params.} & \textbf{FID$\downarrow$} & \textbf{IS$\uparrow$} & \textbf{Pre.$\uparrow$} & \textbf{Rec.$\uparrow$} \\
\midrule
\multirow{3}{*}{\textbf{GAN}} & BigGAN~\citep{brock2018large} & 112M & 6.95 & 224.5 & 0.89 & 0.38 \\
 & GigaGAN~\citep{kang2023scaling} & 569M & 3.45 & 225.5 & 0.84 & 0.61 \\
 & StyleGanXL~\citep{sauer2022stylegan} & 166M & 2.30 & 265.1 & 0.78 & 0.53 \\
\midrule
\multirow{5}{*}{\textbf{Diffusion}} 
% & ADM (Dhariwal and Nichol 2021a) & 554M & 10.94 & 101.0 & 0.69 & 0.63 \\
 & LDM-4 \citep{rombach2022high} & 400M & 3.60 & 247.7 & - & - \\
 & DiT-L/2 \citep{peebles2023scalable} & 458M & 5.02 & 167.2 & 0.75 & 0.57 \\
 & DiT-XL/2 & 675M & 2.27 & 278.2 & 0.83 & 0.57 \\
& \rebut{SiT-XL/2 (ODE)~\citep{ma2024sit}} & \rebut{675M} & \rebut{2.15} & \rebut{258.1} & \rebut{0.81} & \rebut{0.60} \\
 & \rebut{SiT-XL/2 (SDE)} & \rebut{675M} & \rebut{2.06} & \rebut{277.5} & \rebut{0.83} & \rebut{0.59} \\
\midrule
\multirow{2}{*}{\textbf{Mask.}} & MaskGIT \citep{chang2022maskgit} & 227M & 6.18 & 182.1 & 0.8 & 0.51 \\
 & MaskGIT-re & 227M & 4.02 & 355.6 & - & - \\
\midrule
\multirow{7}{*}{\textbf{AR}} & VQGAN \citep{esser2021taming} & 227M & 18.65 & 80.4 & 0.78 & 0.26 \\
 & VQGAN & 1.4B  & 15.78 & 74.3 & - & - \\
 & VQGAN-re & 1.4B & 5.20 & 280.3 & - & - \\
 & ViT-VQGAN \citep{yu2022vector} & 1.7B & 4.17 & 175.1 & - & - \\
 & ViT-VQGAN-re & 1.7B & 3.04 & 227.4 & - & - \\
 & RQTran. \citep{lee2022autoregressive} & 3.8B & 7.55 & 134.0 & - & - \\
 & RQTran.-re & 3.8B & 3.80 & 323.7 & - & - \\
\midrule
\multirow{4}{*}{\textbf{VAR}} & VAR-d16 \citep{tian2024visual} & 310M & 3.30 & 274.4 & 0.84 & 0.51 \\
 & VAR-d20 & 600M & 2.57 & 302.6 & 0.83 & 0.56 \\
 & VAR-d24 & 1.0B & 2.09 & 312.9 & 0.82 & 0.59 \\
 & VAR-d30 & 2.0B & 1.92 & 323.1 & 0.82 & 0.59 \\
\midrule
 \multirow{3}{*}{\textbf{MAR}} & MAR-B \citep{li2024autoregressive} & 208M & 2.31 & 281.7 & 0.82 & 0.57 \\
 & MAR-L & 479M & 1.78 & 296.0 & 0.81 & 0.60 \\
 & MAR-H & 943M & 1.55 & 303.7 & 0.81 & 0.62\\
\midrule
\multirow{6}{*}{\textbf{AR}} & LlamaGen-B \citep{sun2024autoregressive} & 111M & 5.46 & 193.6 & 0.83 & 0.45 \\
 & LlamaGen-L & 343M & 3.81 & 248.3 & 0.83 & 0.52 \\
 & LlamaGen-XL & 775M & 3.39 & 227.1 & 0.81 & 0.54 \\
 & LlamaGen-XXL & 1.4B & 3.09 & 253.6 & 0.83 & 0.53 \\
 & LlamaGen-3B & 3.1B & 3.05 & 222.3 & 0.80 & 0.58 \\
 \midrule
\multirow{3}{*}{\textbf{Ours}} & \modelname-L-d24 & 336M & 2.71 & 275.7 & 0.84 & 0.53 \\
& \modelname-XL-d24 & 799M & 2.49 & 278.8 & 0.84 & 0.55 \\
& \modelname-XXL-d24 & 1.5B & 2.36 & 277.2 & 0.83 & 0.55 \\
\bottomrule
\end{tabular}}
\end{table}

\begin{table}[t]
\centering
\caption{\textbf{Linear-probe top-1 accuracy on ImageNet-1K.} MIM denotes masked image modeling. $^\dag$: our evaluation results.}
\label{appx:compare-linprobe}
\scalebox{0.95}{
\begin{tabular}{ll|cc|c}
\toprule
& \textbf{Method} & \textbf{\#Tokens} & \textbf{Params} & \textbf{ACC1$\uparrow$} \\
\midrule
\multirow{8}{*}{\rotatebox[origin=c]{90}{\textbf{Contrastive methods}}} 
& CPC v2~\citep{henaff2020data} & - & 303M & 71.5 \\
& MoCo~\citep{he2020moco} & - & 375M & 68.6 \\
& SimCLR~\citep{chen2020simclr} & - & 375M & 76.5 \\
& SwAV~\citep{caron2020swav} & - & 93M & 75.3 \\
& DINO~\citep{caron2021dino} & - & 85M & 75.3 \\
& BYOL~\citep{grill2020byol} & - & 375M & 78.6 \\
& CAE~\citep{chen2024context} & - & 304M & 78.1 \\
& CMAE~\citep{huang2023contrastive} & - & 86M & 73.9 \\
\midrule
\multirow{4}{*}{\rotatebox[origin=c]{90}{\textbf{MIM}}}
& iBOT~\citep{zhou2022ibot} & - & 304M & 81.0 \\
& BEiT~\citep{bao2022beit} & 16$\times$16 & 307M & 73.5 \\
& MAE~\citep{he2022masked} & 14$\times$14 & 304M & 75.8 \\
& MAGE~\citep{li2023mage} & 16$\times$16 & 328M & 78.9 \\
\midrule
\multirow{9}{*}{\rotatebox[origin=c]{90}{\textbf{Generative methods}}} 
& BiGAN~\cite{donahue2017adversarial} & - & 138M & 31.0 \\
& BigBiGAN~\citep{donahue2019large} & - & 86M & 56.6 \\
& BigBiGAN & - & 344M & 61.3 \\
& iGPT-L~\citep{chen2020generative} & 32$\times$32 & 1.4B & 60.3 \\
& iGPT-L & 48$\times$48 & 1.4B & 65.2 \\
& ViT-VQGAN-B~\citep{yu2022vector} & 32$\times$32 & 650M & 65.1 \\
& ViT-VQGAN-L & 32$\times$32 & 1.7B & 73.2 \\
& RCG~\citep{li2023self} & 16$\times$16 & 304M & 77.6 \\
& \emph{l}-DAE~\citep{chen2024deconstructing} & - & 304M & 75.0 \\
\midrule
\multirow{6}{*}{\rotatebox[origin=c]{90}{\textbf{Cond. gen.}}} 
& LlamaGen-L$^\dag$~\citep{sun2024autoregressive} & 16$\times$16 & 343M & 40.5 \\
& MAR-B$^\dag$~\citep{li2024autoregressive} & 16$\times$16 & 208M & 57.9 \\
& MAR-L$^\dag$ & 16$\times$16 & 479M & 59.1 \\
& MAR-H$^\dag$ & 16$\times$16 & 943M & 60.0 \\
& \modelname-L-d20 (Ours) & 16$\times$16 & 336M & 67.5 \\
& \modelname-XL-d32 (Ours) & 16$\times$16 & 799M & 69.8 \\
% & \modelname-XXL-d20 (Ours) & 16$\times$16 & \\
\bottomrule
\end{tabular}}
\end{table}

\end{document}